\pdfoutput=1

\documentclass[11pt]{article}
\usepackage{booktabs}
\usepackage{graphicx}
\usepackage{subfig}
\usepackage{caption}
\usepackage{booktabs}
\usepackage{longtable}
\usepackage[]{acl}
\usepackage{graphicx} 
\usepackage{makecell}
\usepackage{amsmath}
\usepackage{amssymb}
\usepackage{bbding}
\usepackage{pifont}

\usepackage{times}
\usepackage{latexsym}
\usepackage{multirow}
\usepackage{subcaption}
\usepackage{mathtools}
\usepackage{verbatim}
\usepackage{adjustbox}
\usepackage{enumitem}
\usepackage{colortbl}
\usepackage{latexsym}
\usepackage{amsfonts}
\usepackage{amssymb}
\usepackage{booktabs}
\usepackage{array}
\usepackage{enumerate}
\usepackage{enumitem}
\usepackage{paralist}
\usepackage[T1]{fontenc}
\usepackage{hyperref}

\usepackage[utf8]{inputenc}
\usepackage{tabularx}
\usepackage{microtype}
\usepackage{algorithm}
\usepackage{algorithmic}
\usepackage{fancyvrb}
\usepackage{makecell}
\usepackage{fdsymbol}
\title{A Multi-Perspective Analysis of Memorization in Large Language Models }

\author{Bowen Chen, Namgi Han, Yusuke Miyao \\
  Department of Computer Science,  The University of Tokyo \\
  \texttt{\{bwchen, hng88, yusuke\}@is.s.u-tokyo.ac.jp} \\
  } 

\begin{document}
\maketitle
\begin{abstract}
Memorization, meaning that Large Language Models (LLMs) can generate content used to train them, is one of the special behaviors of LLMs.
Through studies by previous research, little attention has been paid to what makes sentences memorized, how model size affects it, and the dynamics of generating it.
In this study, we discuss memorization from multiple perspectives, including scaling model size, input and output dynamics, and less unmemorized content, and reveal:
(1) The inter-correlation between memorized/unmemorized sentences, model size, continuation size, and context size, as well as the transition dynamics between sentences of different memorization scores.
(2) A boundary effect when generating memorized/unmemorized content and its relation to model size.
(3) Sentences of different memorization scores cluster in the embedding space are observed through an embedding dynamics analysis.
Entropy analysis in decoding also shows a boundary effect when generating memorized/unmemorized content, albeit oppositely.
(4) The possibility of predicting memorization and its relation to model size and continuation length.
Further analysis showed that unmemorized sequences are easier to predict than memorized content, which can be explained by the significance of the boundary effect.
\end{abstract}

\section{Introduction}
Large language models (LLMs), spanning from BERT \cite{devlin2019bert} to the  GPT-4 \cite{openai2024gpt4}, have revolutionized the Natural Language Processing (NLP) and the artificial intelligence research field.
Though surprised by their performance on various tasks, we still know little about the underlying mechanism of LLMs whose structures are based on neural models known as black-box \cite{alain2018understanding}.
Additionally, due to its unprecedented model size and pre-train data size, LLMs start to present unique behaviors  \cite{wei2022emergent}, and one of them is \textit{memorization}.

Memorization \cite{hartmann2023sok} in the context of LLMs means the LLM can generate the same content recorded in their pre-train corpus under certain contexts. 
On the one hand, utilizing memorization, we can use LLMs as a knowledge base \cite{petroni2019language} that can be served for multiple uses.
On the other hand, personal information contained in the pre-train corpora may also be elicited maliciously \cite{Yao_2024}.
Previous research \cite{tirumala2022memorization, carlini2023quantifying, biderman2023pythia} have studied memorization at the macro level, leaving more micro yet important questions left under-explored, e.g., what makes sentences memorized, what role does the model size play, the input and output dynamics while generating memorized or unmemorized content and how it is related to less memorized content.
To answer the above questions, we conducted a multi-perspective study relating memorization to factors like  the scaling model size, the input and output dynamics, general statistics, and the prediction of memorization, and we found that:
\begin{asparaenum}[(I)]
\item For both memorized and unmemorized sentences, the increase or decrease in model size follows a non-linear trend, indicating a maximum capacity for memorization. 
    The number of memorized sentences decreases sub-linearly with continuation size and increases super-linearly with context size. Moreover, transitions between sentences of different memorization scores were observed, showing that only limited sentences will be memorized with the increase in model size while most sentences remain unmemorized.
    \item We conducted an input dynamics study focusing on the n-gram frequency of inputs. 
    A boundary effect was observed when the model began generating memorized/unmemorized content: the frequency of generating unmemorized tokens suddenly decreased while increased for memorized tokens. 
    This boundary effect is also observed at the sentence level and is more obvious in smaller models, indicating that the boundary effect's significance determines the ease of memorization for a sentence.
    \item Additionally, we analyzed decoding dynamics by examining entropy over vocabulary and the drift of decoded embeddings. 
    Entropy analysis revealed an inverse boundary effect, where entropy suddenly increases for unmemorized content and decreases for memorized content. 
    The embedding dynamics analysis showed sentences with different memorization scores cluster in the embedding space, where the mutual embedding distance grows  with model size
    The close distance between sentences with close memorization scores also suggests the existence of paraphrased memorization. 
    \item Finally, we trained a Transformer model to predict memorization based on context, showing the possibility of predicting memorization.
    Additionally, the results suggest that predicting memorization is easier in large models and easier when predicting unmemorized content, which can be explained by the significance of the boundary effect.
\end{asparaenum}

\section{Related Works}

\subsection{Scaling Laws of LLM}
In this study, our experiments span across various model sizes, thus relating to the research of Scaling Laws \cite{kaplan2020scaling, abnar2021exploring, epoch2023scalinglawsliteraturereview}, which suggests the LLM's performance scales with the size of corpora, the parameter size and the computation required.

Those researches inspired LLMs to continuously scale those factors to gain higher performance like T5 \cite{raffel2023exploring}, GPT-3 \cite{NEURIPS2020_1457c0d6}, PaLM 2 \cite{anil2023palm}, etc.
On the other hand, researchers also analyzed how the scaling affects particular tasks rather than general performance.
For example, \citet{sun-miceli-barone-2024-scaling} discussed how scaling affects translation tasks and prompt injection attacks. 
Within those discussions, the LLM's emergent abilities \cite{wei2022emergent} are also discovered, which means the LLM suddenly reaches high performance on previous low-performance tasks when the LLM reaches a specific model size.
Recent studies have questioned whether emergent abilities are just mirages caused by misuse of metrics \cite{schaeffer2023are} or whether emergent abilities are just context-learning \cite{lu2023emergent}.

Regarding scaling in the field of memorization, \citet{carlini2023quantifying} discussed the memorization across model size and found that the number of memorized texts grows with the model size and context size.
Additionally, \citet{biderman2023emergent} also discussed similar topics and found that a large portion of memorized text in a small-size model is also memorized by a larger model, showing that memorized texts may share certain common features.

\subsection{Memorization}
Before LLM and in real-world practices,  over-fitting is close to memorization  \cite{pub.1055400446}, which means a near-zero loss in the train set that suggests the neural model perfectly memorized the relation between input and data label \cite{zhang2017understanding}.
However, memorization is different from over-fitting as LLM has good performance in the real world while overfitting is usually accompanied by low performance in the test set.
\citet{10.1145/3357713.3384290} has conducted a theoretical analysis in classification models explaining why memorization is actually needed. 
They showed that in a long-tail distribution where many categories only have a few samples, the neural model cannot extract general features from them. 
The best choice for the neural model is just to memorize them and compare them with data in the test set. 
\begin{figure*}[t] 
\centering 
\includegraphics[width=0.75\textwidth]{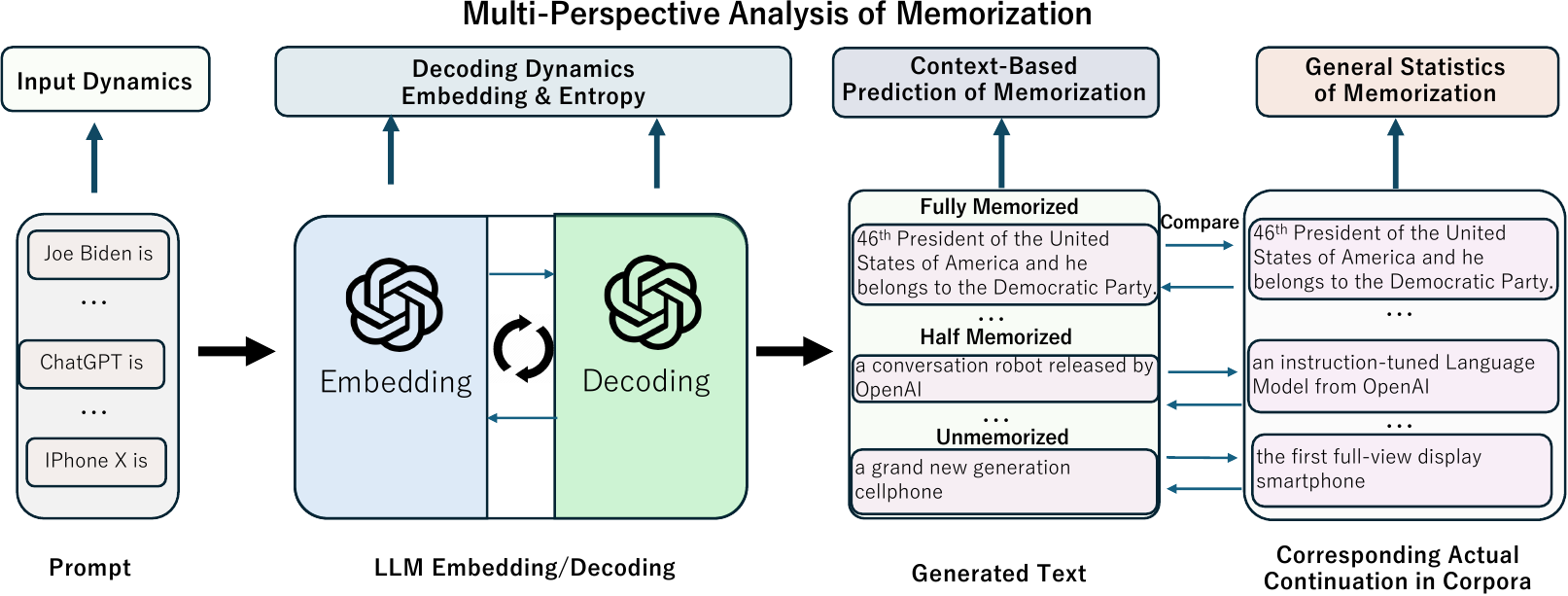} 
\caption{Memorization and Research Scope in this study}
\label{fig: memorization and research scope}
\end{figure*}

For LLMs, different from classification models, they directly generate their pre-train content making the memorization directly observable. which in one way can be used to form a knowledge graph \cite{petroni2019language, alkhamissi2022review} but in another way also leads to the concern of data contamination \cite{sainz-etal-2023-nlp} and privacy risk \cite{Yao_2024}.
Previous research has discussed memorization from a macro level, \citet{tirumala2022memorization} showing large models tend to memorize samples more easily in the training phase.
\citet{carlini2023quantifying} discussed memorization behavior in LLM regarding their factors of model size, continuation size, and context size, showing their inter-correlations.
\citet{biderman2023emergent} studied memorization in LLM regarding their training phase and the overlap between different model sizes, showing the existence of commonly memorized sentences.

Compared to the previous memorization studies, our research focuses on more micro and under-explored questions, e.g., the dynamics while generating them, the reason why some sentences are memorized, and the prediction of memorization.

\section{Experiment Seeting}
\subsection{Criteria}
\subsubsection{Memorization Criteria}
We choose the K-extractability \cite{274574} to define the memorization and calculate the memorization score based on it.

We prompt the LLM with a sequence of context tokens $C= \{c_0 \cdots c_n \}$ whose length is defined as context length and use greedy decoding to generate the continuation tokens based on the prompted context.
Then, we compare the predicted continuations with the actual continuations and calculate a memorization score using the following equation:
\begin{equation}
M(X, Y)=\dfrac{\sum_{i=0}^{n} x_i == y_i}{L(Y)}
\end{equation}
In the above equation, $X= \{ x_0 \cdots x_n \}$ is the sequence of predicted continuation tokens, and $Y= \{ y_0 \cdots y_n \}$ is the true continuation tokens.
$L(Y)$ means the length of the true continuation tokens.
If $M(X, Y) == 1$, this sequence is fully memorized under this context sequence, called as \textit{K-extractable}. 
Similarly, a sequence $Y$ in unmemorized if $M(X, y)==0$.\footnote{Due to the number of continuation tokens, the memorization score has different granularities.
In some experiments, we classify sentences into several memorization levels based on their memorization scores.}

\subsubsection{Prediction Criteria}
This study uses two criteria to evaluate model performance in predicting memorization, e.g., Token-Level Accuracy and Full Accuracy.

The prediction for a sequence of generated tokens is $\hat{X}=\{\hat{x_0},\hat{x_1} \dots \hat{x_n} \}$ where each prediction $\hat{x_n}$ is a binary label indicating the token at this index is a memorized token or not.
Similarly, the gold result is denoted as $\hat{Y}=\{\hat{y_0},\hat{y_0} \dots \hat{y_n}\}$ where each $\hat{y}$ is the golden label.

The Token Level Accuracy can be defined as:
\begin{equation}
T(\hat{X}, \hat{Y}) = \dfrac{\sum_{i=0}^{n} \hat{x_i} == \hat{y_i}}{L(\hat{Y})}   
\end{equation}
The Full Accuracy means that the prediction of token level at a sentence must be entirely correct, which means $T(\hat{X}, \hat{Y}) == 1$.

\subsection{Model Setting}
We use the Pythia model \cite{biderman2023pythia} to analyze the memorization behavior as Pythia provides LLMs trained across various sizes with the same training order using open-sourced Pile \cite{pile} corpora, which benefits the experiment and provides stability.

Specifically, we investigate the model size of [70m, 160m, 410m, 1b, 2.8b, 6.9b, 12b] where $m$ stands for million and $b$ stands for billion.
We choose the model trained on the deduplicated pre-train data version to avoid the effect of duplicated sentences, as previous research reports that the chance to be memorized grows exponentially with the number of duplicates \cite{kandpal2022deduplicating}.\footnote{For more details, please refer to \url{https://github.com/EleutherAI/pythia}}
\begin{figure*}[t] 
\centering 
\includegraphics[width=1\textwidth]{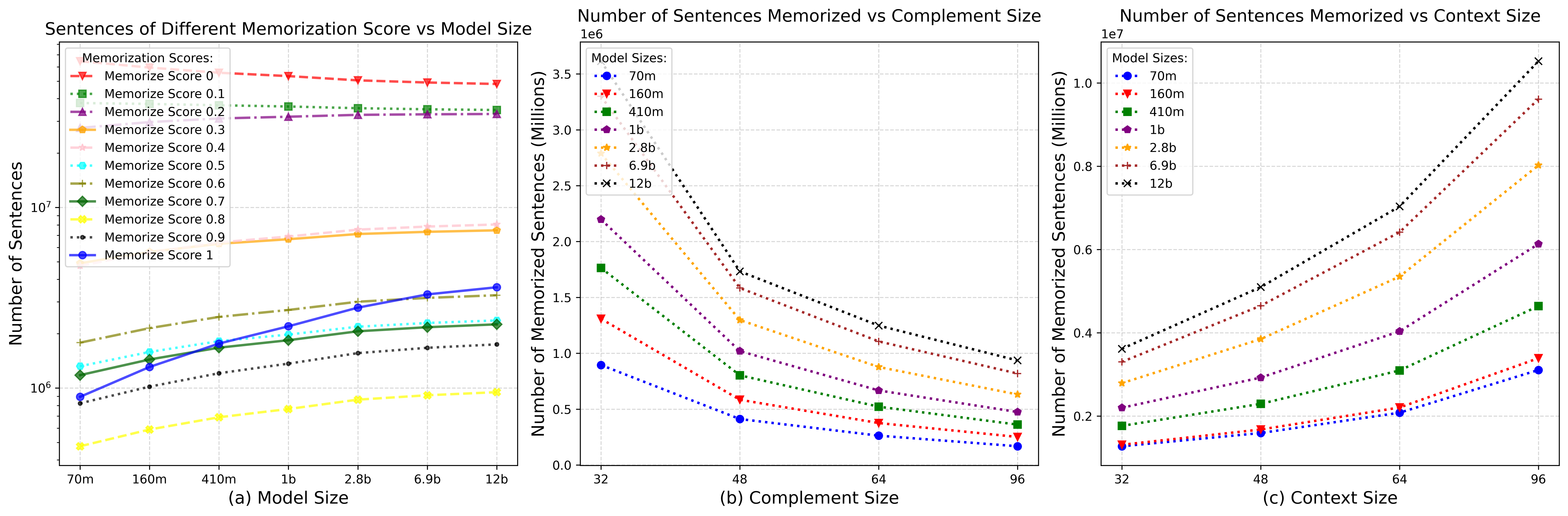} 
\caption{Memorization Statistics Across Model Size, Complement Size, and Contenxt Size}
\label{fig: memorization and model size}
\end{figure*}
\section{Experiment Results}
\subsection{Memorization Factors}
This section discusses how memorization is affected by different factors.
We collect the number of sentences memorized across different extend under different model sizes.
Then, we divide them into ten sets with a memorization difference of 0.1.
The results are plotted in Figure \ref{fig: memorization and model size}.

Firstly, we can see that the memorized sentences increase with the model size and context size but decrease with the increase of complement size, which aligns with previous research \cite{carlini2023quantifying}.
We illustrate a more in-depth analysis of those factors in the following sections.
\subsubsection{The Factor of Model Size}
In this experiment, we discuss how model size affects the number of memorized and unmemorized contexts.
From the Figure \ref{fig: memorization and model size} (a), we can obtain that:
\begin{asparaenum}[(I)]
 \item The number of sentences with low memorization scores (0-0.3) is significantly higher than those with high memorization scores, indicating that most of the pre-train data are not memorized despite the existence of memorization in LLMs.
    \item Among sentences with high memorization scores, the count of fully memorized sentences increases more rapidly, suggesting that LLMs have a propensity to fully memorize sentences rather than partially. Additionally, the number of sentences with low memorization scores (0, 0.1) decreases as the model size increases, indicating that unmemorized sentences gradually become memorized with larger models.
    \item The increase or decrease in the number of memorized or unmemorized sentences is not linear with respect to model size. 
    There is a noticeable increase in numbers for fully memorized sentences from 70 million to 2.8 billion parameters, compared to 2.8 billion to 12 billion parameters.
    A similar decreasing trend for unmemorized sentences is observed. This suggests a capacity for memorization, implying LLMs cannot memorize the entire corpus even with sufficiently large model sizes.
\end{asparaenum}

\subsubsection{Context and Complement Size}
We change the number of prompted tokens while fixing the complement or context size and then observe the number of fully memorized sequences.
From the Figure \ref{fig: memorization and model size} (b) and (c), we can tell that:
\begin{asparaenum}[(I)]   
    \item The decrease in the number of memorized sentences with increasing complement size is not linear. 
    For instance, the complement size increase from 64 to 96 results in a relatively minor decrease compared to the change from 32 to 48, indicating some sentences are firmly memorized.
    \item The reduction in memorized sentences with increased complement size is more obvious in larger models. 
    This demonstrates that although larger models memorize more sentences, their memorization is less robust compared to smaller models.
    \item The increase in memorized sentences with context size is also non-linear, with longer context leading to an almost exponential rise in the number of memorized sentences. 
    This increase is more significant in larger models, indicating that more contents are potentially memorized in large models, which can be elicited by giving longer context.
\end{asparaenum}

\begin{figure}[t] 
\centering 
\includegraphics[width=1\columnwidth]{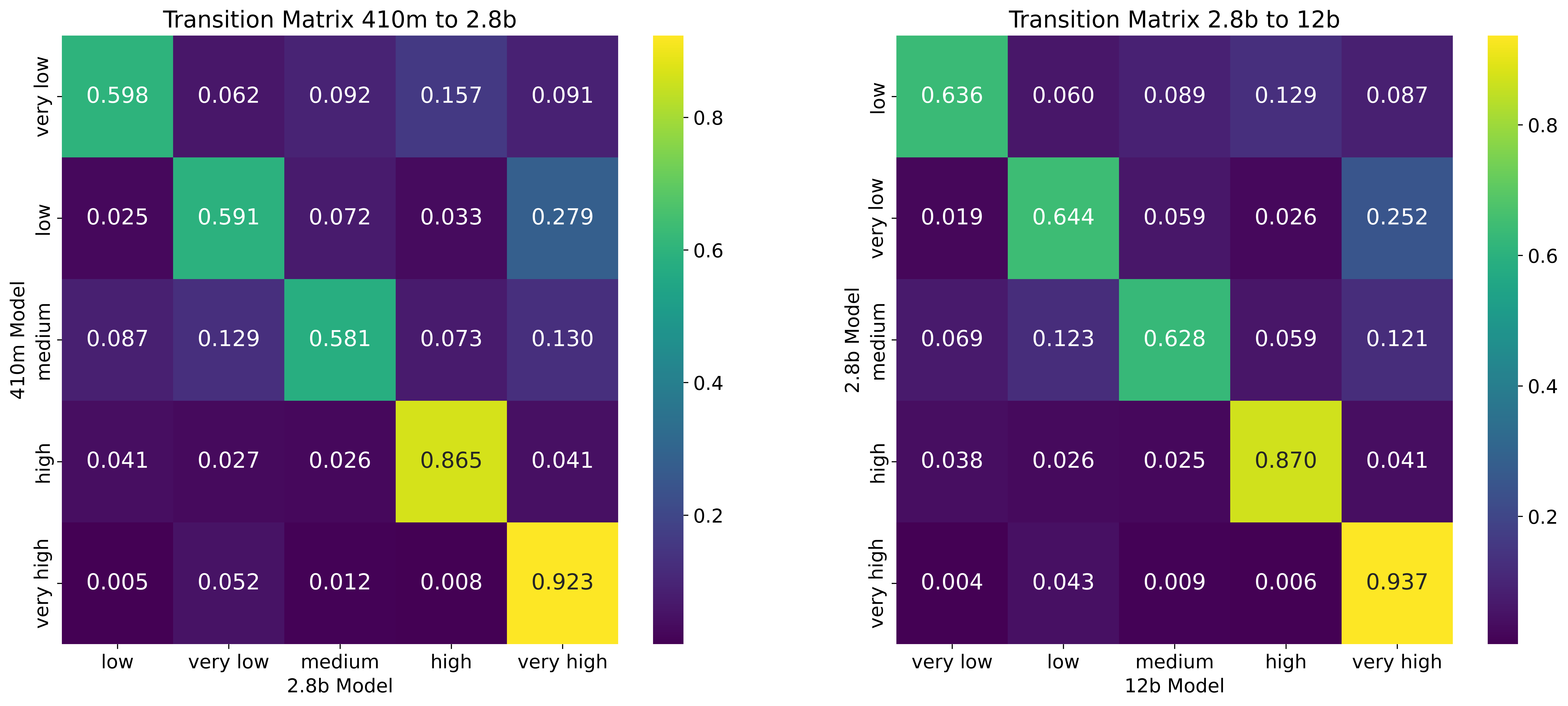} 
\caption{Transition Across Different Model Size}
\label{fig: transition}
\end{figure}

\subsection{Memorization Transition}
This section discusses how sentences with different memorization scores transit across model sizes. 
Specifically, when trained with a larger model, we study the mutual transition between sentences with different memorization scores.
We classify sentences with memorization scores with a range difference of 0.2 with labels of very low, low, medium, high, and very high.
We plot the transition matrix in Figure \ref{fig: transition}, which shows the transition from the 410m size model to the 2.8b size and 2.8b size to the 12b size model, and we can conclude:
\begin{asparaenum}[(I)]   
    \item Most sentences remain in their previous state even when trained with a larger model, as indicated by the diagonal entries in the transition matrices. 
    Additionally, the higher the memorization score, the less likely the sentence will transition to another state.
    For highly memorized sentences, over 90\% remain memorized compared to those with low memorization scores.
    \item With increasing model size, sentences are more likely to stay in their original state. 
    For example, the transition probability at the diagonal is higher when moving from the 2.8 billion to the 12 billion parameter model compared to the transition from the 410 million to the 2.8 billion parameter model. 
    This suggests memorized or unmemorized states become more fixed as the model size increases.
    \item Even for highly memorized sentences, there is a small chance of transitioning to a low memorized state, implying that some sentences may be memorized randomly rather than due to specific features. 
    This randomness allows for the possibility of transitioning to a low memorized state when trained on a larger model.
\end{asparaenum}

\subsection{Input Dyanamics}
In this section, we discuss the question of \textit{whether there is any sign when the model starts to generate memorized or unmemorized content.}
Especially what makes sentences memorized at different extents and why some sentences are memorized by large-size models but not small-size models.
\subsubsection{Token Level Frequency Analysis}
\begin{figure}[t] 
\centering 
\includegraphics[width=1\columnwidth]{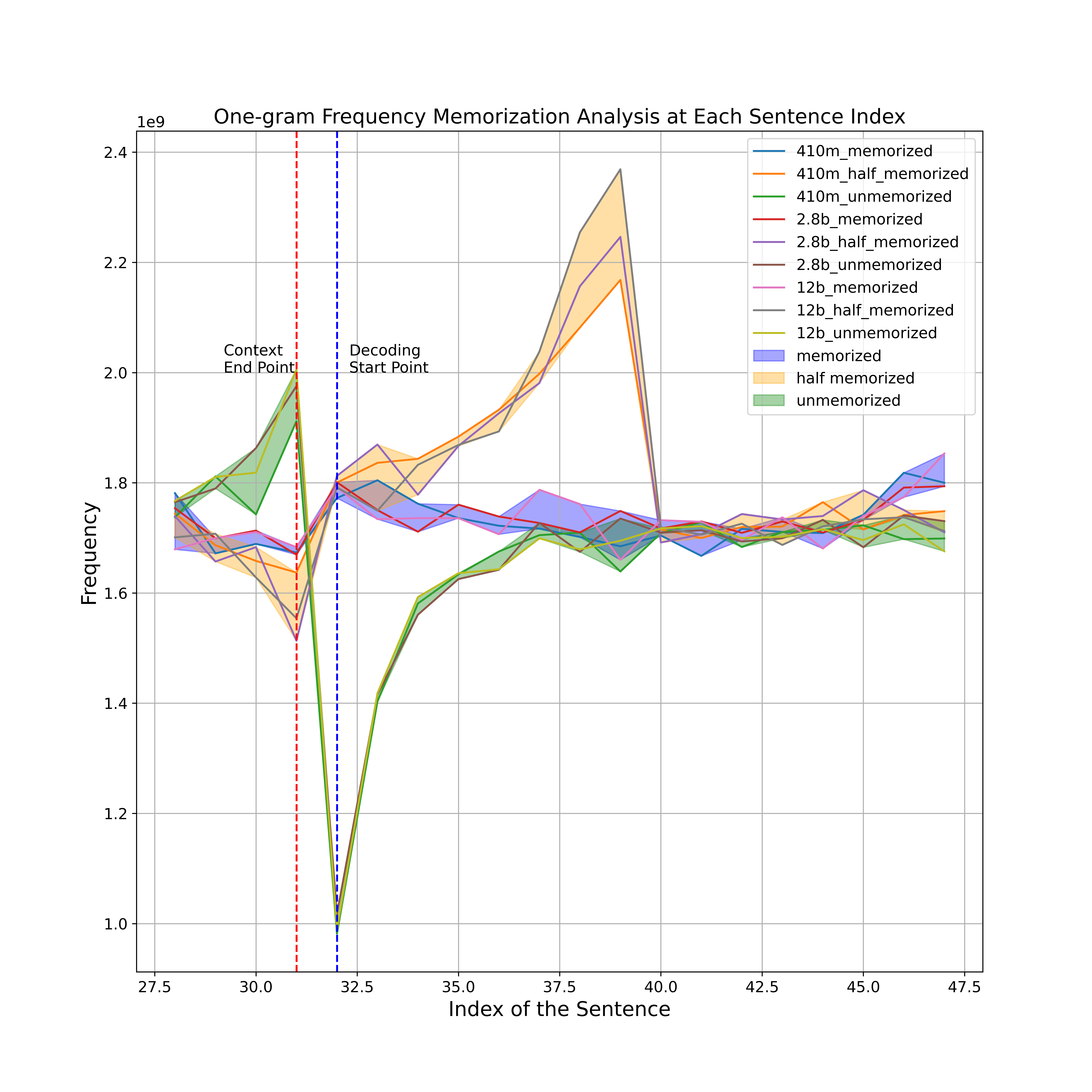} 
\caption{One-gram Analysis at Each Index}
\label{fig: n-gram}
\end{figure}
Firstly, we begin with the input level by analyzing the n-gram statistics of the pre-train corpora.  
We show the 1-gram statistics across steps in Figure  \ref{fig: n-gram}, and we can see:
\begin{asparaenum}[(I)]  
    \item A clear boundary effect is observed. Around index 32, representing the first generated token, the frequency drops and then rises for memorized sentences (\textit{positive boundary effect}), whereas it rises and then drops for unmemorized sentences (\textit{negative boundary effect}). 
    The negative boundary effect is more pronounced than the positive one. 
    \item For the half-memorized sentence, we see the frequency increase and drop ((\textit{negative boundary effect})) around the index of 39, which is half the length of generated tokens.
    This shows that for the half-memorized sentence, the previous half is mostly memorized, and the later half is mostly unmemorized, meaning that the memorized tokens are distributed in a near continuous way rather than scattered in the generated tokens.
    \item The positive boundary effect in memorized sentences suggests that memorization is driven by the higher frequency of initial tokens, implying that remembering the first few tokens makes the entire sentence easier to memorize.
    Conversely, the negative boundary effect in unmemorized sentences indicates that the low frequency of initial tokens makes the following sequences easier to forget.
\end{asparaenum}
\subsubsection{Sentence Level Frequency Analysis }
Given the existence of the boundary effect at the token level, we discuss how such an effect exists across model sizes and whether other factors influence memorization at the sequence level.
We calculate the average frequency of context and continuation tokens and frequency difference at the boundary in Table \ref{tab: gram results}, and we can obtain:
\begin{asparaenum}[(I)]   
    \item The frequency of one-grams is significantly higher than that of two-grams, approximately 3.5 times higher on average in both context and continuation.
    The boundary effect is consistent in the two-gram setting, though the positive boundary effect is less obvious due to the frequency drop when computed with two-grams. Despite this, the actual frequency gap remains substantial (million level) when considering the unit is billion.
    \item In memorized sentences, the frequency is lower in the context and higher in the continuation, whereas unmemorized sentences exhibit the opposite pattern. 
    This suggests that the boundary effect persists at a broader sequence level, although it is less obvious.
    \item For half-memorized sentences, the frequency of continuation tokens is higher than the context average frequency.
    This is due to the frequency increase before reaching the first generated unmemorized tokens, as indicated in Figure \ref{fig: n-gram}.
    \item In the Boundary Frequency Difference column, we see a decreasing trend in both positive and negative boundary effects.
    However, the decrease in the positive boundary effect makes it less significant, while the decrease in the negative boundary effect makes it more significant.
    These results imply that the significance of the positive boundary effect correlates with the ease of memorizing a sentence. In contrast, the significance of the negative boundary effect correlates with the ease of not memorizing a sentence.
\end{asparaenum}
\begin{table*}[!tb]\scriptsize
    \centering
   \begingroup
    \setlength{\tabcolsep}{4pt}
    \begin{tabular}{@{}ccccccccccccccccccc@{}}
    \toprule
       \multirow{3}{*}{\textbf{Size}}  & \multicolumn{6}{c}{\textbf{Context}} & \multicolumn{6}{c}{\textbf{Continuation}} & \multicolumn{6}{c}{\textbf{Bounrady Freq Difference}} \\
       \cmidrule(lr){2-7}\cmidrule(lr){8-13}
       \cmidrule(lr){14-19}
       &
       \multicolumn{3}{c}{One}&\multicolumn{3}{c}{Two}&\multicolumn{3}{c}{One(B)}&\multicolumn{3}{c}{Two}&\multicolumn{3}{c}{One}&\multicolumn{3}{c}{Two}
       \\
       \cmidrule(lr){2-4}\cmidrule(lr){5-7}\cmidrule(lr){8-10} \cmidrule(lr){11-13}
       \cmidrule(lr){14-16}\cmidrule(lr){17-19}
       & M & H& U& M &H& U& M &H& U& M &H& U& M &H& U& M &H&U\\
       \midrule
       160m & 1.708&1.713&1.744&0.551&0.534&0.691&1.739 &1.837&1.628&0.535&0.659&0.567&0.114& 0.330&-0.939&0.033&0.101&-0.663\\
       \midrule
       1b  &  1.713&1.711&1.752&0.558&0.552&0.697&1.736 &1.832&1.631&0.509&0.682&0.564&0.103&0.270&-0.981&0.028&0.090&-0.696\\
       \midrule
       6.8b& 1.721&1.710&1.759&0.570&0.565&0.701&1.736&1.829&1.638&0.496&0.699&0.564&0.090&0.140&-0.963&0.027&0.085&-0.726\\
       \midrule
       12b& 1.721&1.720&1.760&0.572&0.569&0.702&1.736 &1.846&1.626&0.493&0.704&0.563&0.039&0.237&-1.016&0.026&0.083&-0.732\\
       \bottomrule
    \end{tabular}
    \endgroup
    \caption{Gram Statistics. The frequency unit is billion. Boundary Freq Difference means we use the frequency of the generated token to subtract the frequency of the last token in the context.}
    \label{tab: gram results}
\end{table*}

\subsection{Output Dynamics}
\subsubsection{Embedding Dynamics}
In this section, we discuss the embedding of generated tokens for sentences with different memorization scores. 
We collect the hidden state at the last layer of each generated token for sentences with different memorization scores and compute the pair-wise Euclidean distance and cosine similarity between them.
Then, we perform PCA \cite{doi:10.1080/14786440109462720} on those embeddings and visualize as shown in Figure \ref{fig: embedding dyanamics}, and we can obtain:
\begin{figure}[t] 
\centering 
\includegraphics[width=1\columnwidth]{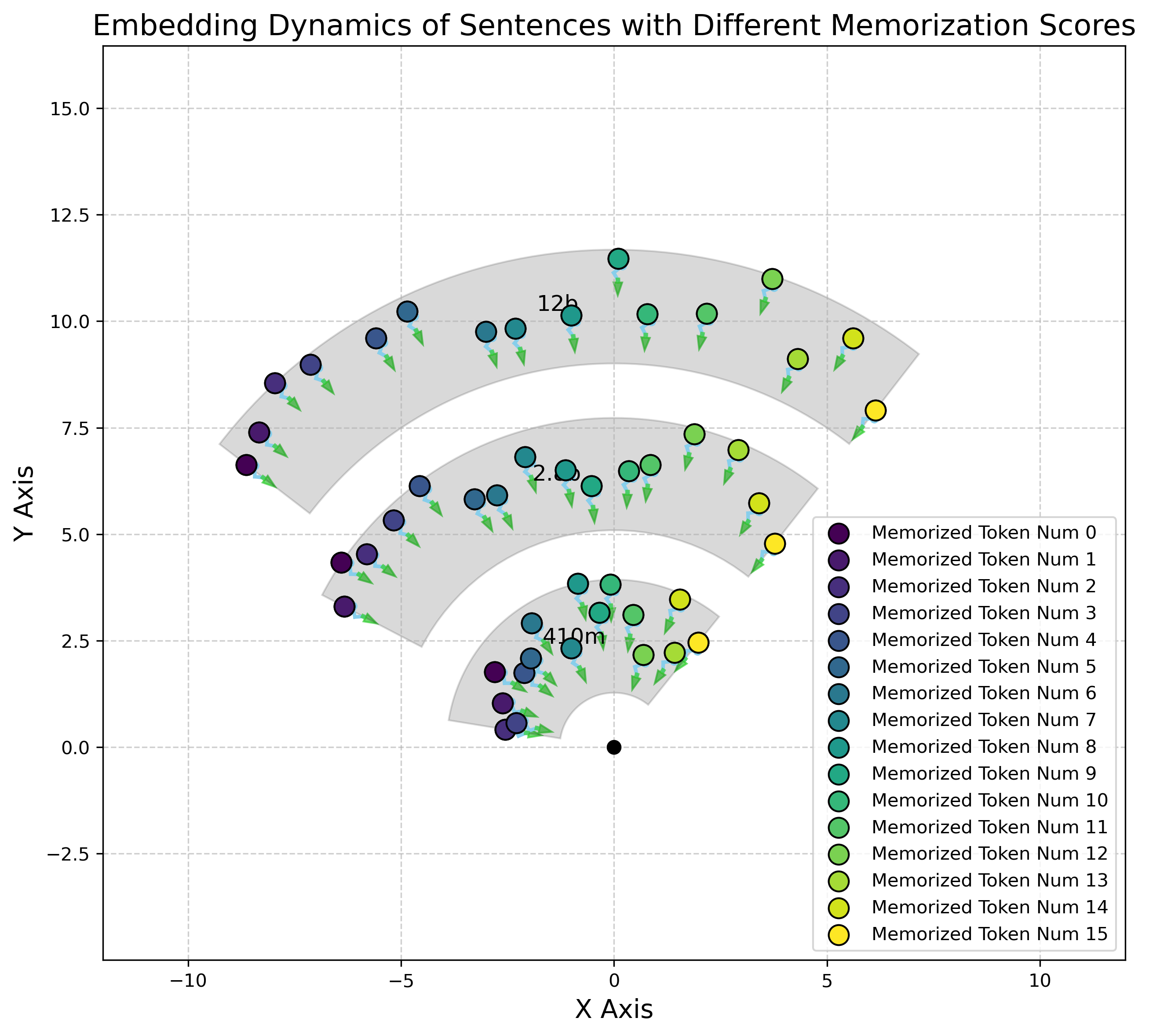} 
\caption{Embedding Dynamics of Across Different Model Size. Memorized Token Num $x$ means $x$ generated tokens are the same with the true continuation.}
\label{fig: embedding dyanamics}
\end{figure}
\footnote{Detailed numbers of Cosine Similarity and Euclidean distance between sentences with different memorization scores are at different decoding steps are at Appendix \ref{embedding dyanamics}}
\begin{asparaenum}[(I)]   
    \item The cosine similarity remains relatively stable across different encoding steps between sentences memorized at different extents. 
    Meanwhile, from the movement of those points, we can see the Euclidean distance decreases with the generation of tokens.
    Combined with the previous observation that sentences memorized with different extend have a stable cosine similarity, we can conclude that with the generation of tokens, the angle between sentence vectors at high dimension space remains stable while the cosine similarity decreases.
    \item For highly memorized sentences, even when not fully memorized, they cluster closely in the embedding space. 
    This suggests the generated content is both semantically and lexically similar to the memorized content, indicating the existence of paraphrased memorization.
    \item Larger models exhibit higher Euclidean distances and lower cosine similarities. 
    The higher Euclidean distance is due to the expansion of hidden sizes (e.g., 512 for 70 million, 2048 for 1 billion parameters). 
    The decrease in cosine similarity is also attributed to the expanded hidden sizes, which increase the expressivity of the embedding space. 
    This expansion enlarges the sentence embeddings, reducing cosine similarity but enhancing the differentiation between sentences. 
    This expanded expressivity in embedding helps to explain the performance gap between different model sizes:
    larger models distribute different sentences more distinctly with fewer embedding overlaps, while smaller models mix embeddings more, leading to ambiguity and degraded performance.
\end{asparaenum}
\subsubsection{Generation Dynamics and Entropy}
\begin{figure}[t] 
\centering 
\includegraphics[width=1\columnwidth]{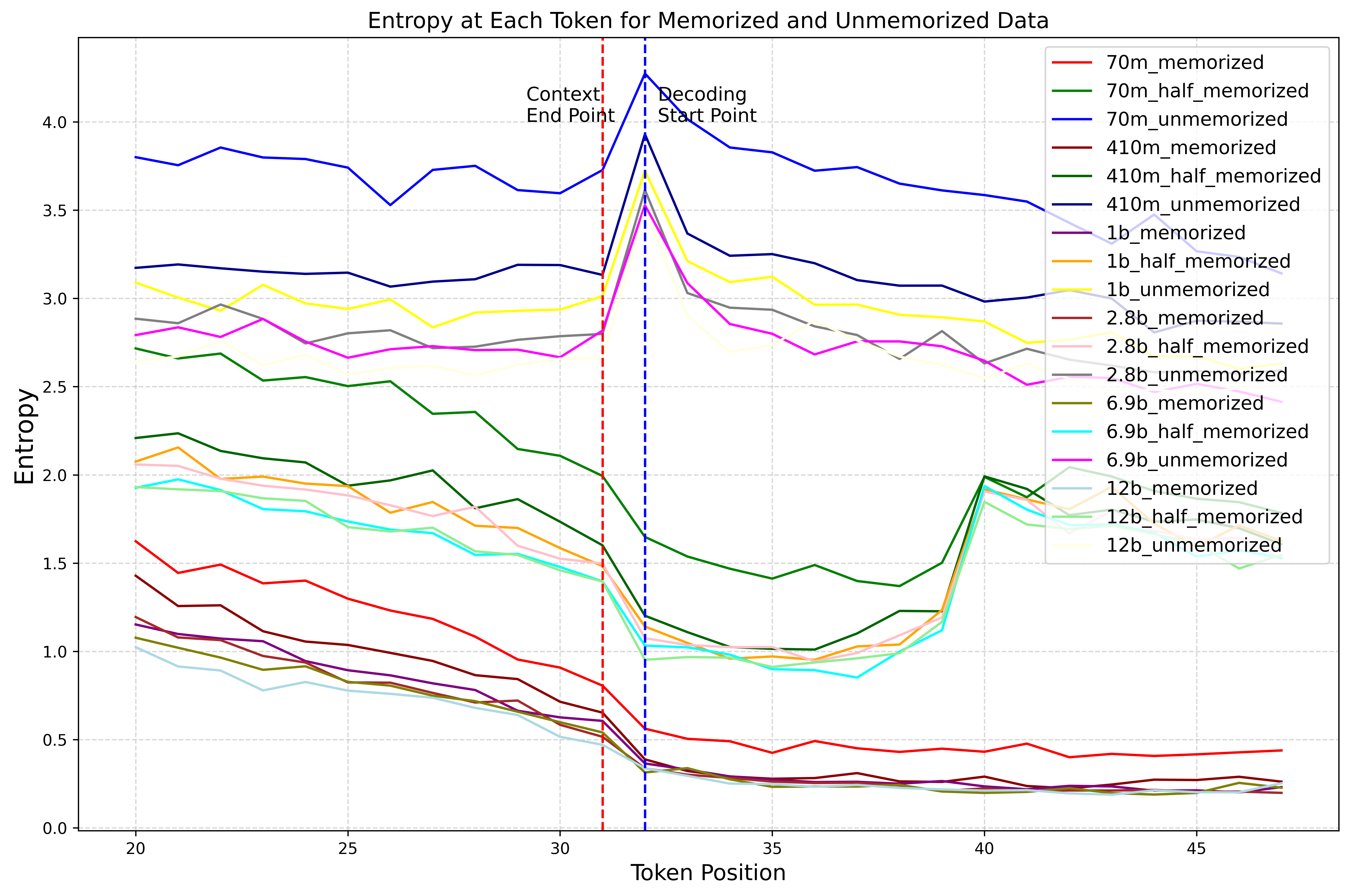} 
\caption{Averaged Entropy at Each Index}
\label{fig: entropy across step}
\end{figure}
This section discusses how the model generates the memorized sentences with different memorization scores, e.g., the generation dynamics.
Given the observed boundary effect at the input level, we aim to address \textit{whether the model exhibits specific behaviors when generating memorized versus unmemorized sentences.}
To investigate this, we collected the generation probability for each token during the generation of memorized, unmemorized, and half-memorized sentences. We then calculated the entropy across 10,000 sentences and plotted the entropy for each token in Figure \ref{fig: entropy across step}. From this analysis, we derive the following observations:
\begin{asparaenum}[(I)]   
    \item First, we can see that the entropy differs depending on the extent of memorization and model size. 
    An unmemorized sentence has a higher average entropy at each token than an unmemorized sentence.
    This shows LLM is more confident when generating memorized content.
    \item Additionally, when the model starts to generate unmemorized tokens, we can see a significant increase in entropy, showing such boundary effect also exists in the decoding steps while behaving oppositely compared to the boundary effect in frequency analysis.
    The entropy drops around the index of 32 for the memorized and half-memorized content.
    This also shows the existence of the boundary effect for memorized content. 
    However, we can see the drop of entropy for the memorized sentence is not samely significant as the half-memorized sentence, showing that even for the memorized content, the model is more confident when the sentence is fully memorized, leading to lower entropy.
    \item We can also see the entropy decrease with the increase in model size. 
    This suggests that larger models are more confident about generating tokens than small ones. 
    Additionally, we can see that the entropy is lower in the context of memorized tokens, showing that the entropy of context also relates to how a sentence is memorized.
    Further, the significance of the boundary effect in entropy decreases with the model size for memorized content while remaining unchanged for unmemorized content.
\end{asparaenum}
\begin{table*}[!tb]\small
\centering
\begin{tabular}{lcccccccccccc}
\toprule
\multirow{2}{*}{\textbf{Length}}&  \multicolumn{2}{c}{\textbf{70m}}&  \multicolumn{2}{c}{\textbf{410m}}&  \multicolumn{2}{c}{\textbf{1b}}& \multicolumn{2}{c}{\textbf{2.8b}}& \multicolumn{2}{c}{\textbf{6.9b}}& \multicolumn{2}{c}{\textbf{12b}}\\
\cmidrule(lr){2-3} \cmidrule(lr){4-5}  \cmidrule(lr){6-7}  \cmidrule(lr){8-9}  \cmidrule(lr){10-11} \cmidrule(lr){12-13}
& Token  & Full & Token  & Full   & Token  & Full  & Token  & Full & Token  & Full & Acc & Token \\

\midrule
16& 78.2 & \underline{10.2}& 78.6 & \underline{10.4} & 78.8 & \underline{10.6} &80.1&\underline{10.7}& 77.4 & \underline{8.3} &80.3&\underline{\textbf{10.9}}\\
\midrule
32& 78.6 & 5.9& 79.6 & 6.0 &79.7&6.1& 80.1&6.3& 80.5&6.4&80.8&6.4\\
\midrule
48& 79.6 & 5.2& 80.3 & 5.4 & 80.4&5.6&80.4&5.5&80.8&5.8&81.0&6.0 \\
\midrule
64& \underline{80.1} & 4.7& \underline{80.8} & 4.8 & \underline{81.2}& 5.2& \underline{81.5}&5.5&\underline{81.8}&5.8&\underline{\textbf{82.1}}&6.0 \\
\bottomrule
\end{tabular}
\caption{Performance of Transformer Model on Prediction of Memorization. The context length is defaulted at 32. Length means the token length that is required to be predicted. Token and Full represent Token-Level Accuracy and Full Accuracy. The best results across model sizes are bold, and the best results within a model size are underlined.}
\label{tab:my_label}
\end{table*}
\subsection{Prediction of Memorization}
In this section, we train a Transformer model to predict the memorization of tokens from the LLM's embedding and statistic features and analyze the results.
We discuss whether Transformer can do this task and the difficulty of predicting the memorized token at different model sizes.
As in the real world and indicated by our experiment results in previous sections, similar context sentences can also trigger memorized texts \cite{stoehr2024localizing}, and the generated token can also be a paraphrase of the actual continuation \cite{ippolito-etal-2023-preventing}.
\subsubsection{Results on Memorization of Prediction}
The training is formalized by receiving both the last layer's embeddings at each step and statistics (e.g., the entropy). 
Then, the Transformer is required to generate a binary classification label at each index, predicting whether the token at this index is memorized.
The results are presented in Table \ref{tab:my_label}.
From the table, we can obtain the following results:
\begin{asparaenum}[(I)]   
    \item First, regarding Token-level accuracy, we can see with a naive Transformer model, the token-level accuracy can reach $80\%$ and even higher, showing that the prediction of memorization at the token level is easy. 
    The full accuracy is low as this requires a correct prediction for every token.
    \item As the model size of LLMs increases, both token-level and full-level accuracy improve.
    This suggests that predicting memorization behavior is easier for larger models because the greater embedding distances make classification easier.
    \item Token-level accuracy increases with the continuation size, likely due to increased training data.
    For instance, a prediction length of 64 involves four times more training data than a prediction length of 16.
    However, full-sentence accuracy decreases as the prediction length increases, making it harder to achieve full accuracy because more tokens need to be predicted correctly.
\end{asparaenum}
\subsubsection{Analysis of Full Accuracy}
\begin{figure}[t] 
\centering 
\includegraphics[width=1\columnwidth]{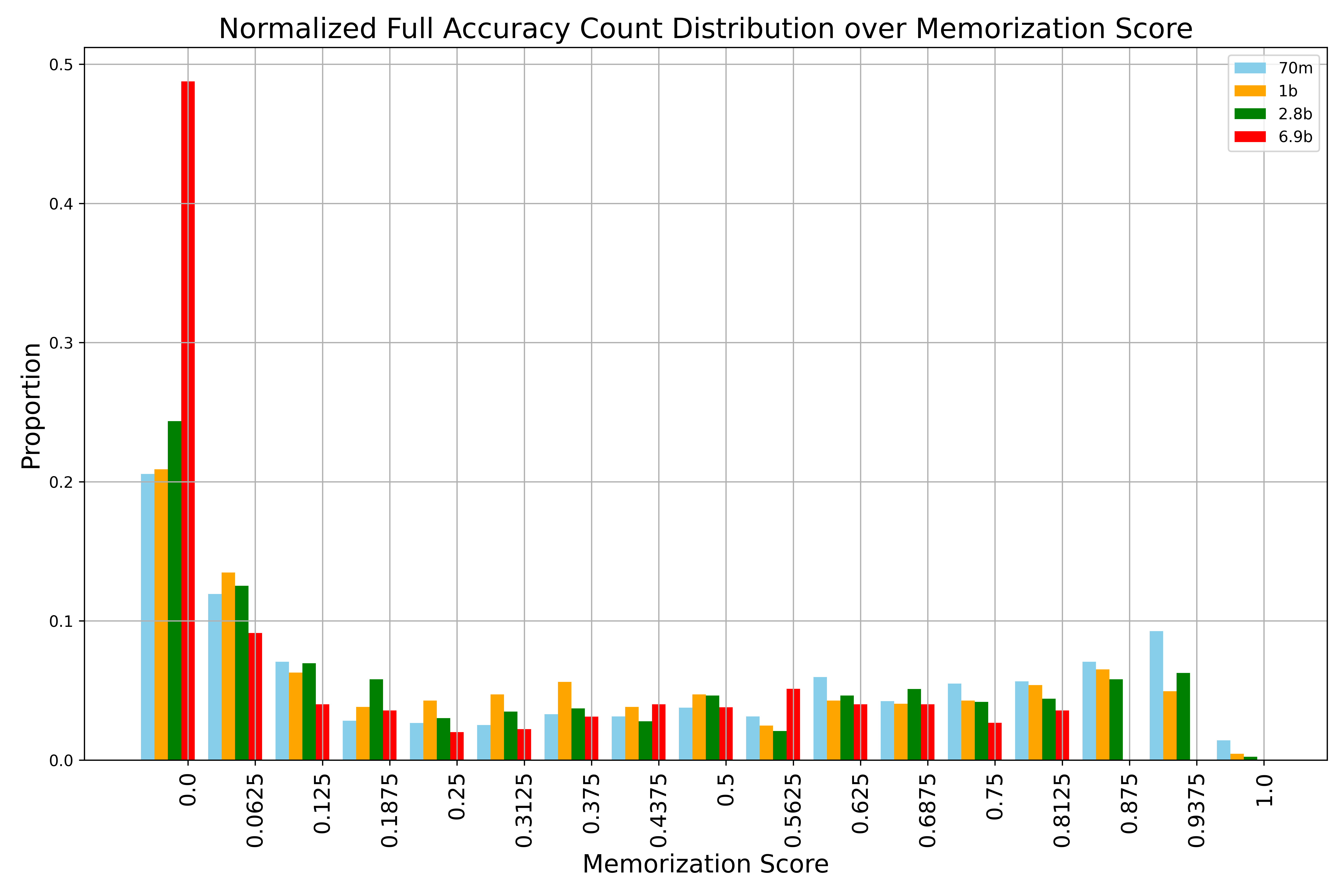} 
\caption{Distribution Across Model Szize of Full Accurate Predictions }
\label{fig:distribution}
\end{figure}
In this experiment, we analyze how fully correct predictions distribute across memorization scores to discuss whether the difficulty in predicting memorization changes with the memorization score.
From the results shown in Figure \ref{fig:distribution}, we can tell:
\begin{asparaenum}[(I)]   
    \item For any model size, they are more good at predicting sentences with low memorization scores. 
    In contrast, the low portion for sentences with high memorization scores indicate they are harder to predict accurately.
    \item As model size increases, the proportion of low memorization scores raise and decrease for high memorization score sentences which even reaches zero for 6.9b model.
    This suggests larger models are more accurate in predicting unmemorized sentences compared to memorized ones.
    \item  The ease of predicting sentences with low memorization scores can be attributed to the significance of the boundary effect.
    Previous experiments show that the boundary effect of unmemorized sequences is more obvious than that of memorized sequences in both token frequency and entropy. 
    Particularly, the boundary effect decreases for memorized sequences while increases for unmemorized sequences with increased model size.
    Such change makes the boundary effect less significant for memorized sentences, making them harder to predict, while more significant for unmemorized sentences, making them easier to predict.
\end{asparaenum}

\subsection{Conclusion}
In this study, we comprehensively discussed LLM memorization from various perspectives.
From a statistical level, our analysis extended the scope of previous research to sentences with lower memorization scores.
This is followed by an experiment showing the memorization transition across model sizes.
At the input dynamic level, through frequency analysis, we found the positive and negative boundary effects when generating memorized and unmemorized tokens and their relation to how easily a sentence can be memorized/unmemorized.
In the output dynamics, at the embedding level, we found clusters of sentences with different memorization scores in the embedding space, and the close distance of sentences with high memorization scores indicates the existence of paraphrase memorization.
At the entropy level, we observed an opposite boundary effect and analyzed its change with model size. 
Finally, we trained a Transformer model to predict the memorization, showing that token-level prediction is easy while sentence-level is challenging.
Through analysis of fully correct predicted samples, we found unmemorized tokens are easier to predict than memorized tokens, which can be explained by the significance of the boundary effect.

\section{Limitations}
This research has analyzed various factors regarding the memorization phenomena of LLMs.
We acknowledge that there are still limitations to this research.
Due to the lack of LLMs whose models and data are both being released, as the data is also different, it is hard to compare the memorization across different LLMs.
Future research can be on how to comprehensively measure memorization of various LLMs, either close-sourced like GPT-4 or open-sourced like Pythia or LLaMa.
Additionally, Pythia only provides LLMs up to 12B, which is still smaller than SotA open-sourced LLMs like LLaMA, whose largest model is 70B.
The difference between 12B  and 70B may lead to emergent abilities, which may also affect the memorization results obtained in 12B, which may not be able to scale even further due to those factors. 
Though we trained the Transformer to predict the memorization of LLMs, it is more analysis-oriented; thus, the performance is not the main focus of this experiment.
Additionally, in this research, the discussion of memorization is under verbatim memorization, where generated tokens are identical to the same sentence in the corpora.

\bibliography{anthology,custom}
\bibliographystyle{acl_natbib}

\appendix

\section{Appendix}
\label{sec:appendix}

\begin{figure*}[htbp] 
\centering
\includegraphics[width=1\textwidth]{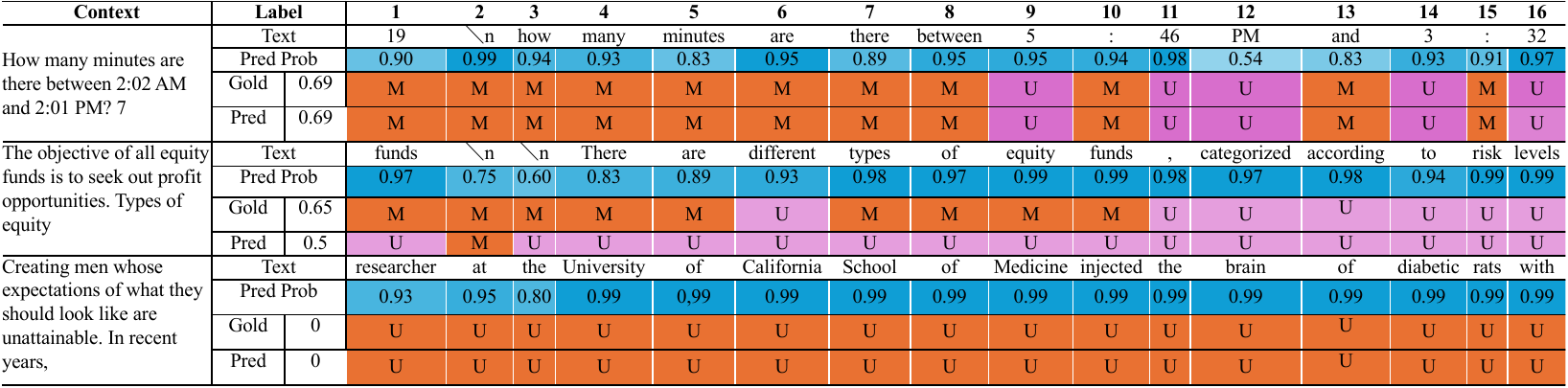} 
\caption{Prediction Examples. Pred Prob means the output predicted probability of the corresponding label in the pred row for each example. Gold means the true label, Pred means the predicted label, and Pred Prob means the probability of the corresponding prediction. $M$ means the label for a token at this index is a memorized token. $U$ means the label for a token at this index is an unmemorized token.}
\label{fig:visulization}
\end{figure*}
\section{Experiment Setting}
The experiment uses 64 A100 40Gbs GPUs when using LLMs to generate tokens given the previous context, which utilizes PyTorch's parallel running packages.
We run the model with half-precision, which increases both the speed and saves memory.
This follows the previous Pythia implementation when generating tokens given context.

The running time depends on the model size and the generated token length.
In the situation when a 70m model with 32 context tokens and 16 tokens is required to be generated, it can be run with one A100 GPU within several hours.
However, if such an experiment is in a single A100 GPU, it would estimated to take two weeks to finish the generation.
Therefore, with 64 GPUs, the running of the 12b model in such a situation can be shortened to around one day.
However, the generation time also largely increases with the length of generated tokens, which grows in a linear way with the number of tokens to be generated.
Additionally, since we use greedy decoding and do not consider other possible decoding options but the token with the highest probability, it is also possible that the running time may be different if using a more complicated decoding strategy.
However, using a more complex decoding strategy does not affect the results.
If a sequence of tokens is memorized, the memorized token will always be the most probable token when generating them.
\subsection{Case Study}
We also provide a case study of the model's prediction on the test set regarding the prediction of memorization in Figure \ref{fig:visulization}.
From this figure, we can see that:
\begin{asparaenum}
    \item Confirming previous experiments, the memorized token is mostly continuous, showing the memorization happens in chunks of sequences rather than individual sequences.
    \item In the first example, the model outputs a fully correct prediction that aligns with the actual label, showing the possibility of predicting memorization by utilizing embedding information.
    \item In the second example, we see that the model's prediction does not align with the actual continuation. The model predicts an unmemorization label for the memorized label.
    \item In the third example, we can see this unmemorized sequence. The model fully predicted those labels. We also noticed that the probability for the corresponding token is very high, showing that the model is very confident about the prediction.
\end{asparaenum}

\begin{figure*}[htbp]
\centering
\includegraphics[width=0.85\textwidth]{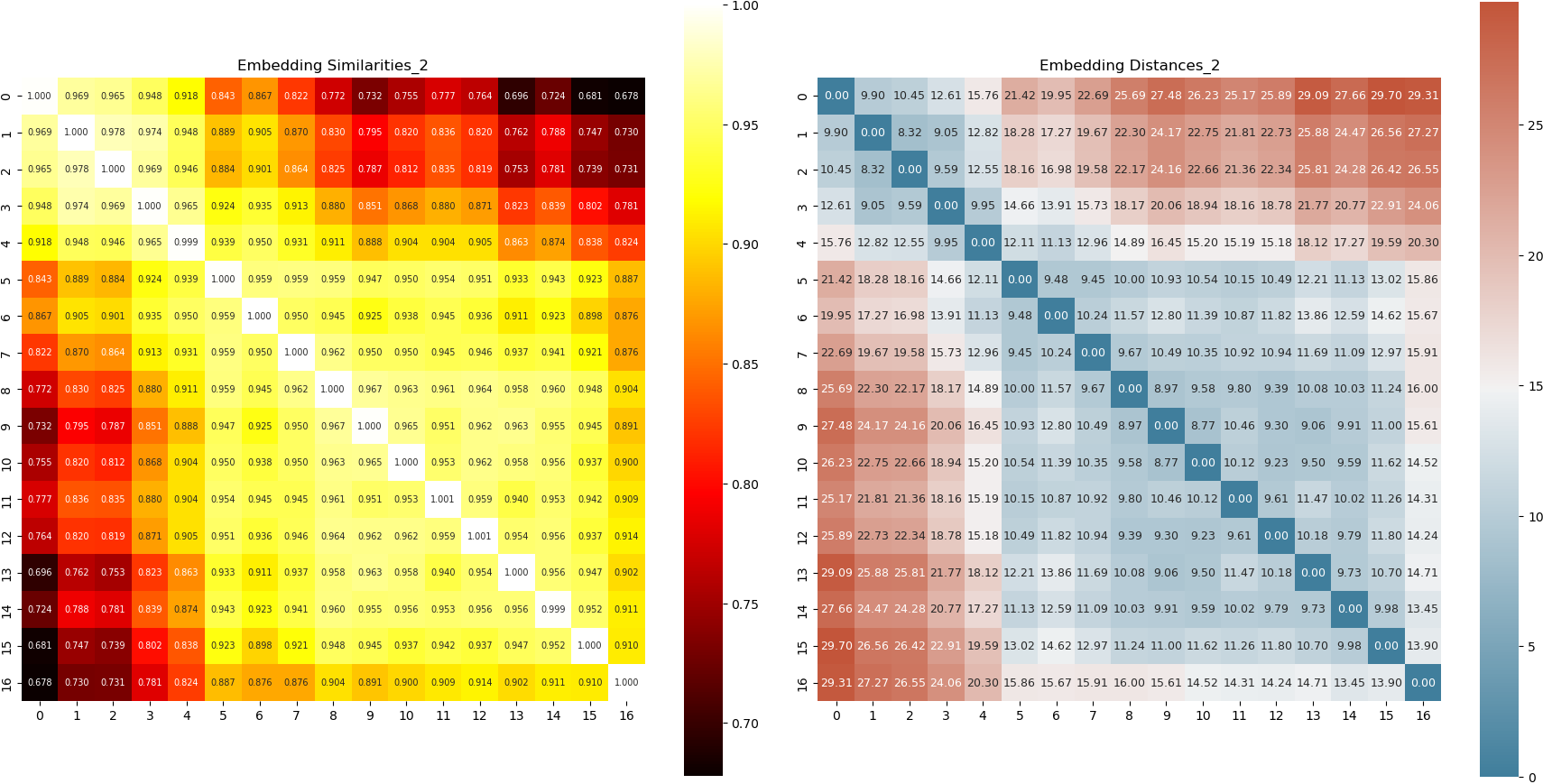}
\caption{410m Model, Step 2}
\label{fig:410m2}
\end{figure*}

\begin{figure*}[htbp]
\centering
\includegraphics[width=0.85\textwidth]{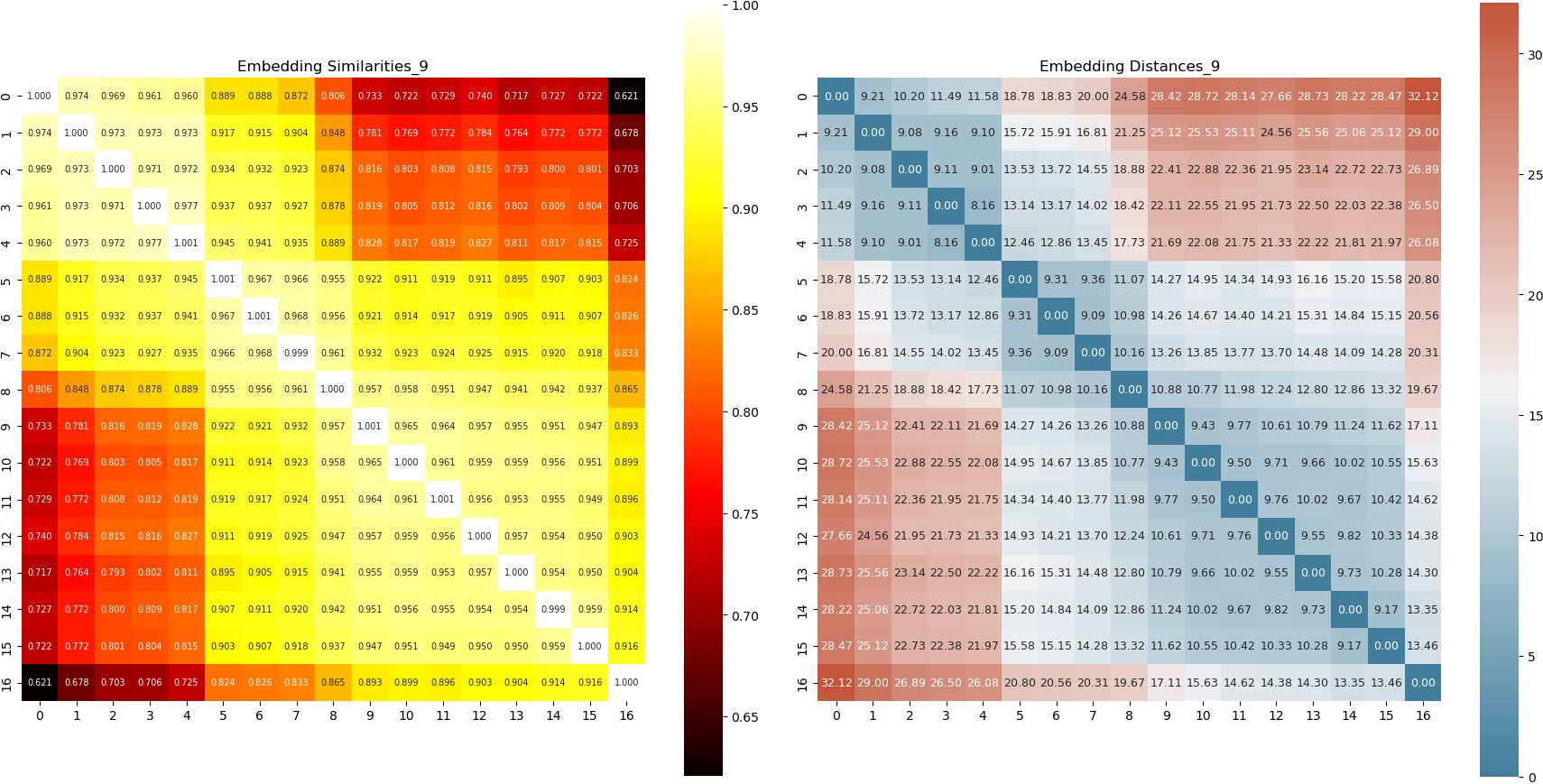}
\caption{410m Model, Step 9}
\label{fig:410m9}
\end{figure*}

\begin{figure*}[htbp]
\centering
\includegraphics[width=0.85\textwidth]{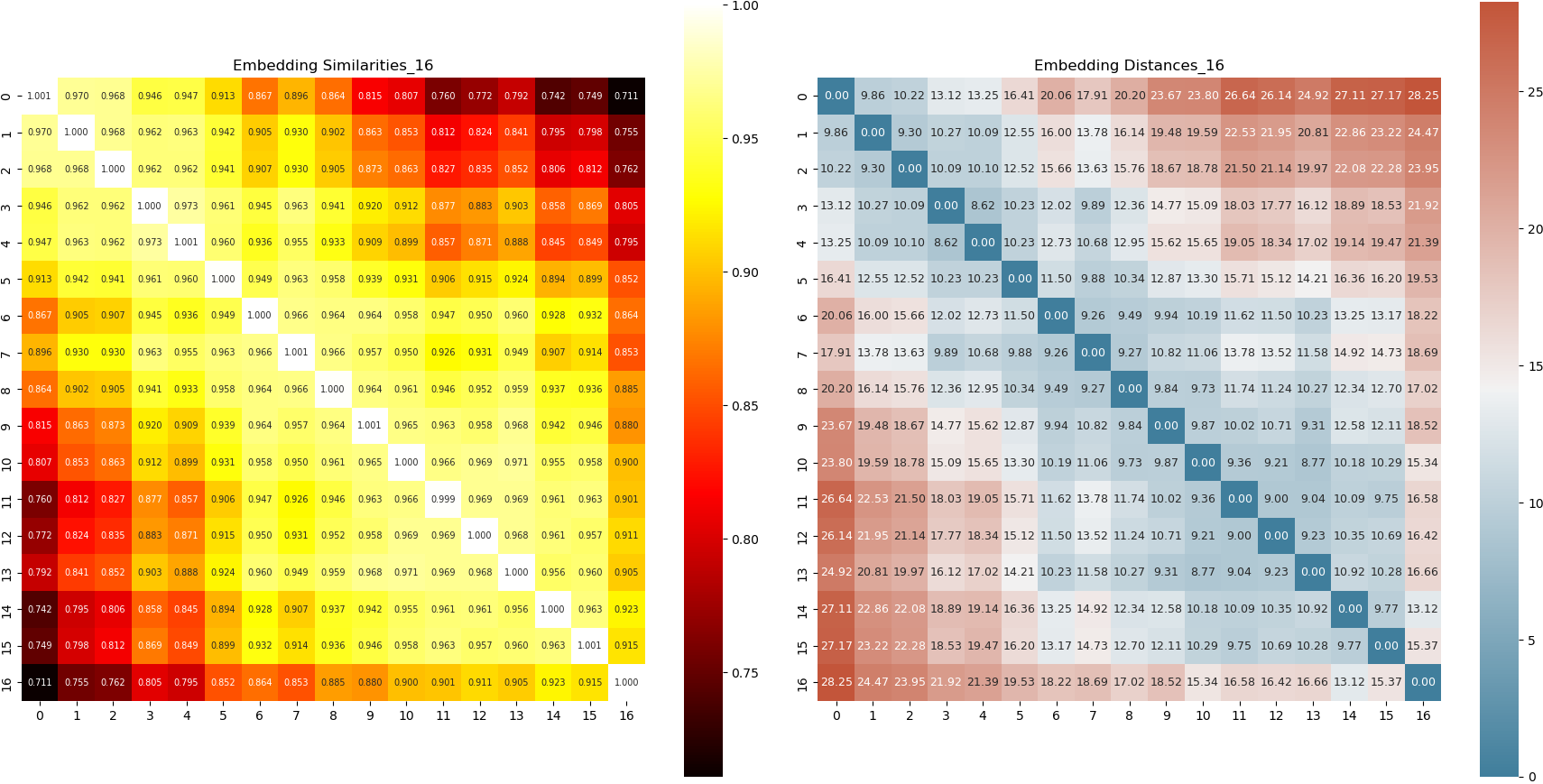}
\caption{410m Model, Step 16}
\label{fig:410m16}
\end{figure*}

\begin{figure*}[htbp]
\centering
\includegraphics[width=0.85\textwidth]{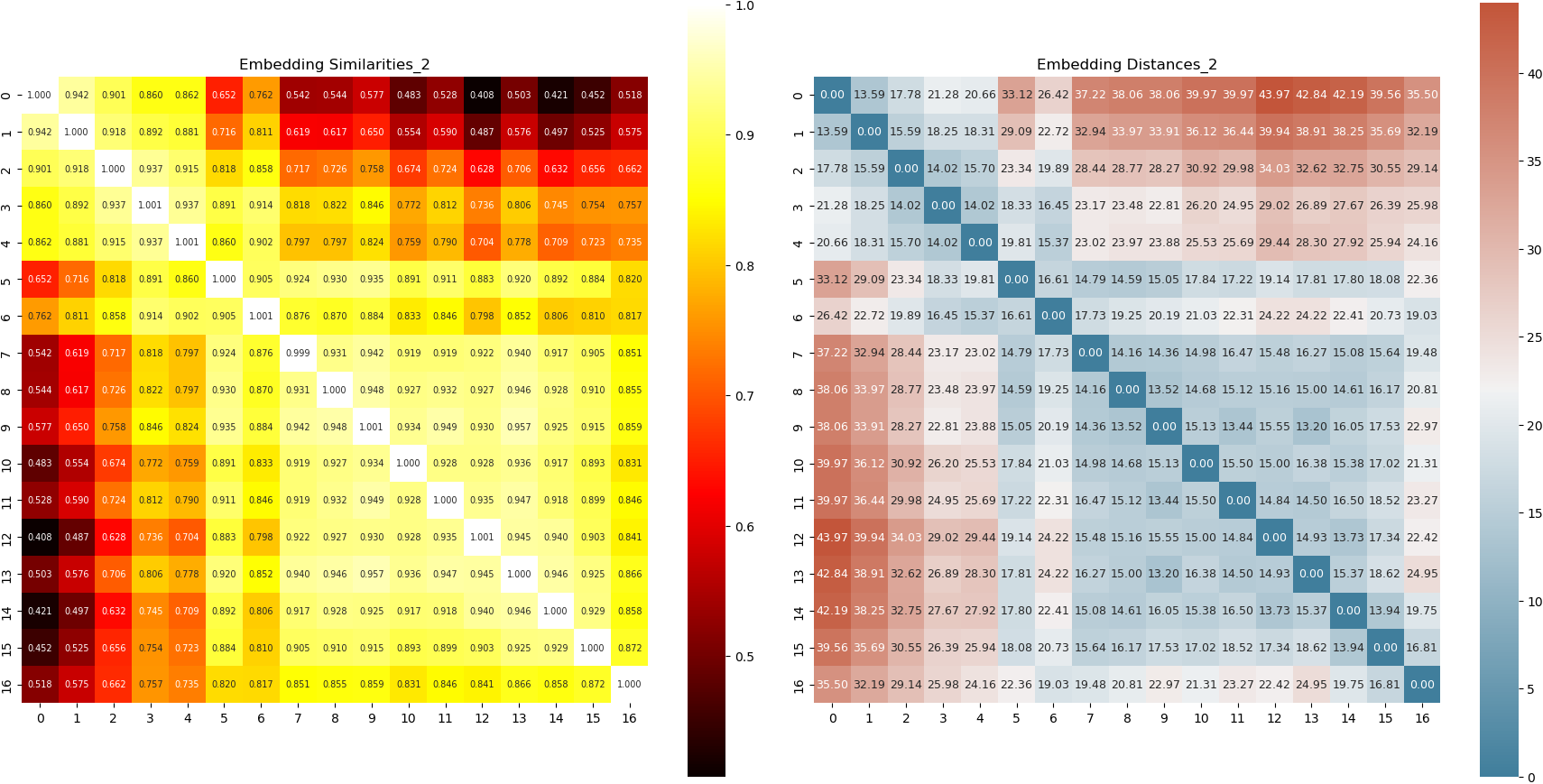}
\caption{2.8b Model, Step 2}
\label{fig:2.8b2}
\end{figure*}

\begin{figure*}[htbp]
\centering
\includegraphics[width=0.85\textwidth]{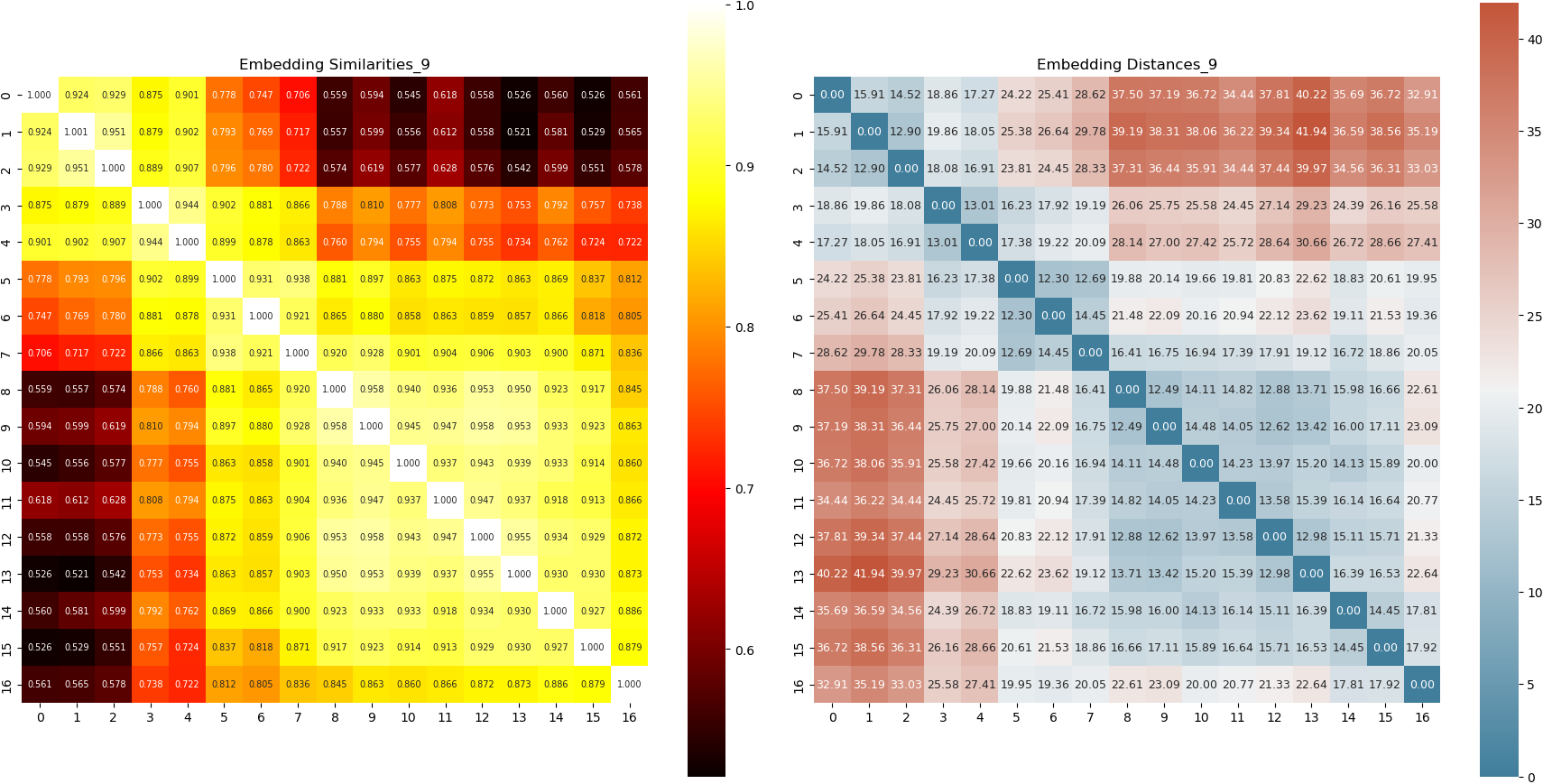}
\caption{2.8b Model, Step 9}
\label{fig:2.8b9}
\end{figure*}

\begin{figure*}[htbp]
\centering
\includegraphics[width=0.85\textwidth]{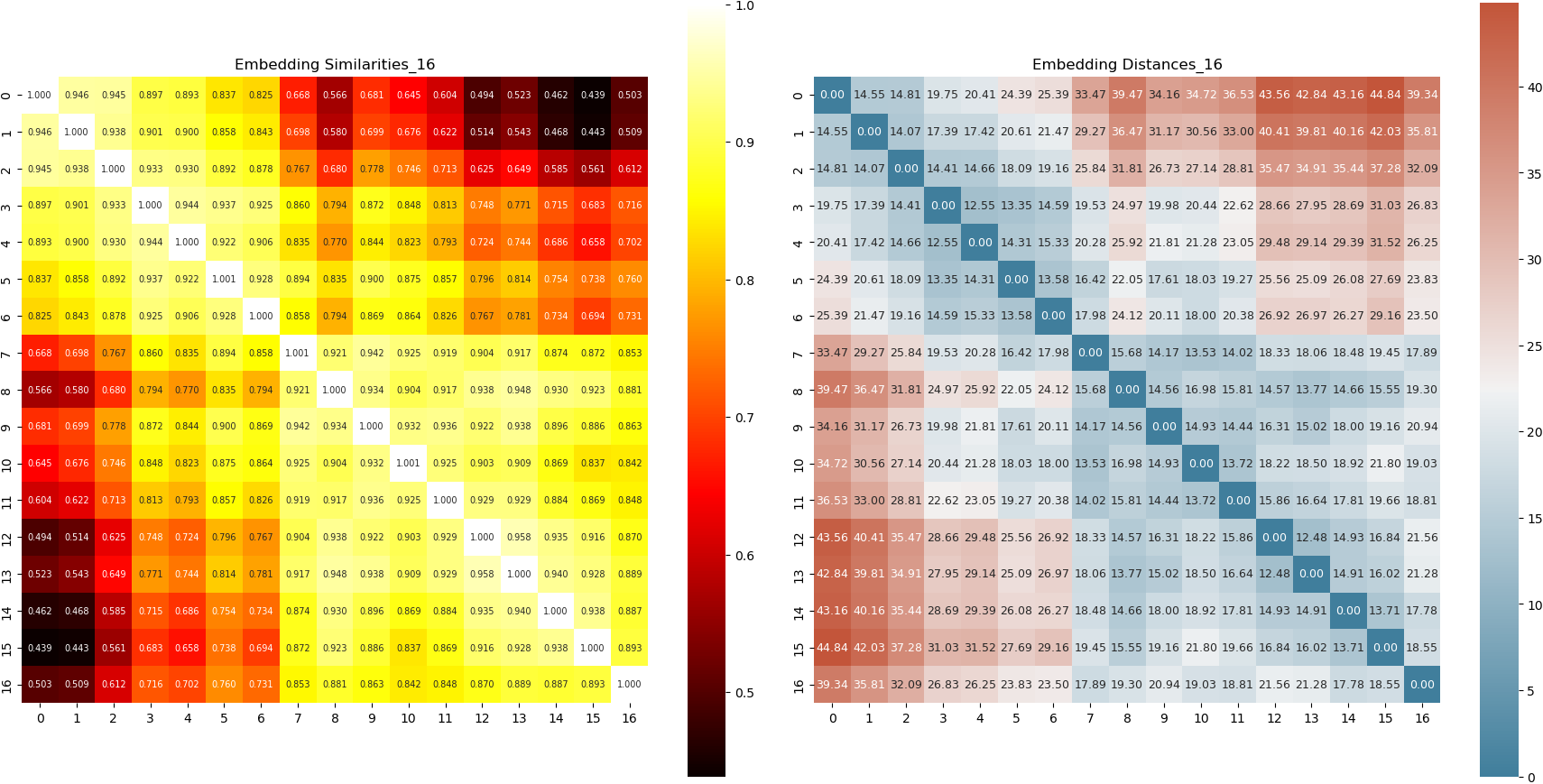}
\caption{2.8b Model, Step 16}
\label{fig:2.8b16}
\end{figure*}

\begin{figure*}[htbp]
\centering
\includegraphics[width=0.85\textwidth]{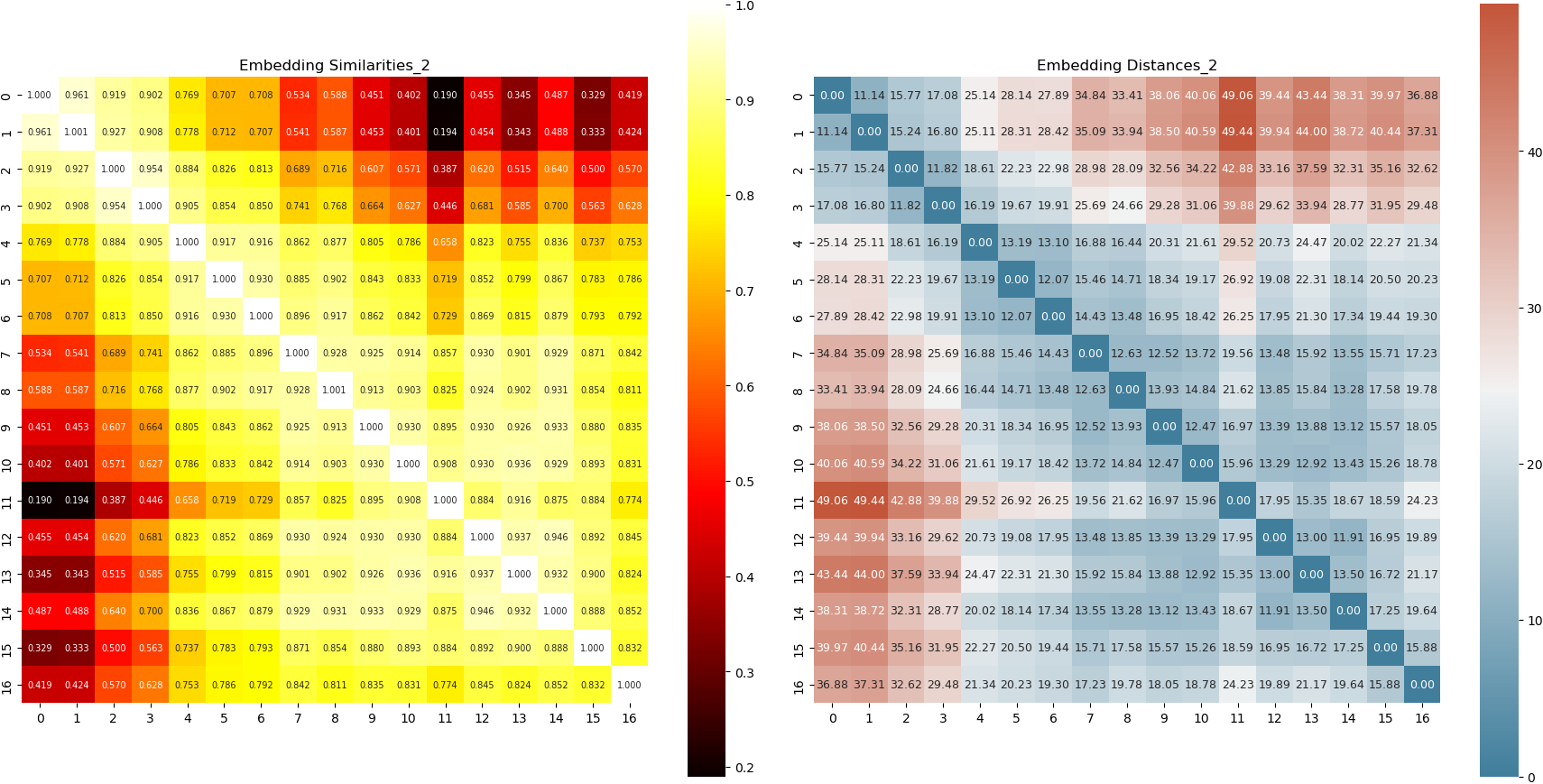}
\caption{12b Model, Step 2}
\label{fig:12b2}
\end{figure*}

\begin{figure*}[htbp]
\centering
\includegraphics[width=0.85\textwidth]{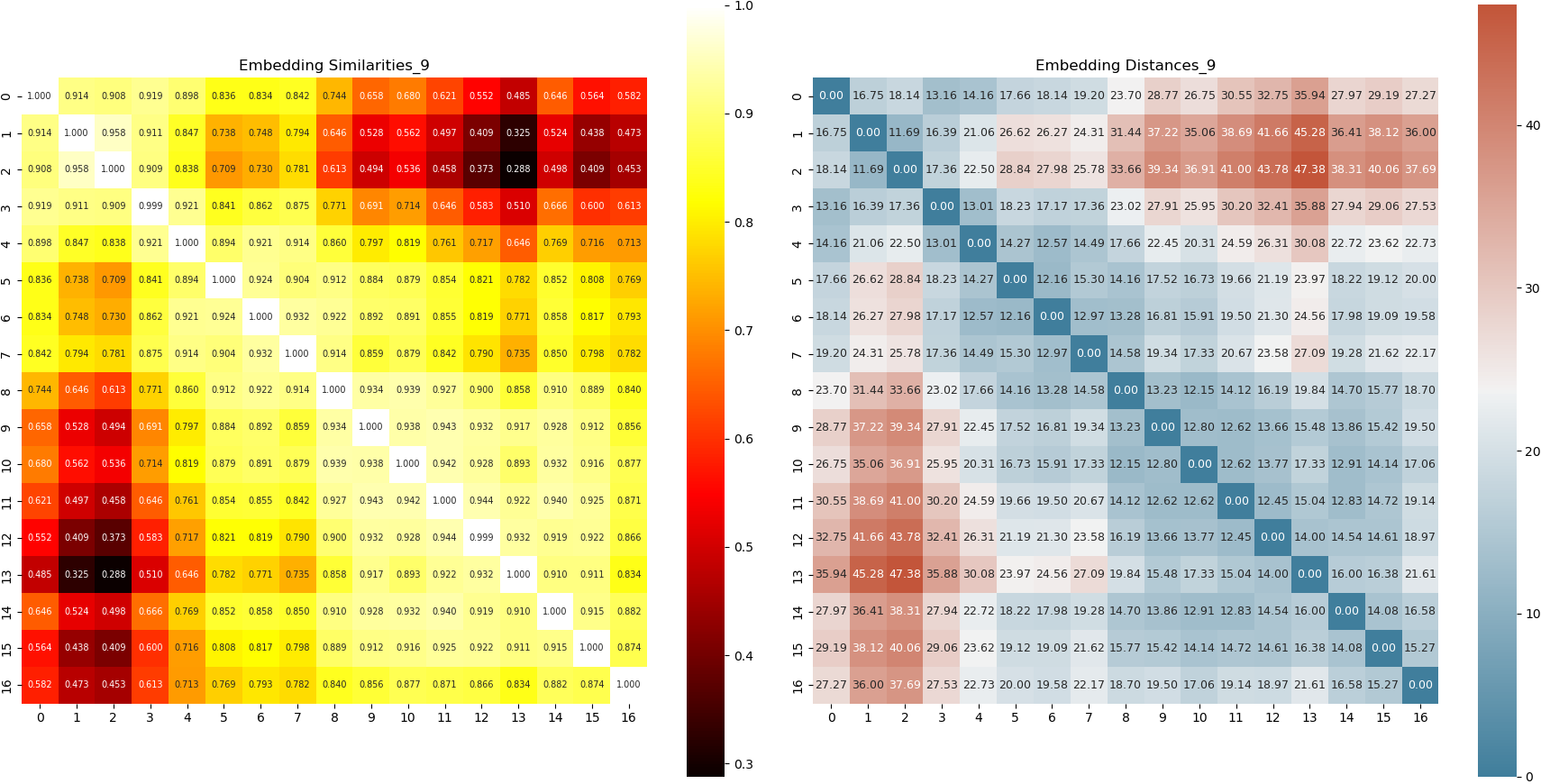}
\caption{12b Model, Step 9}
\label{fig:12b9}
\end{figure*}

\begin{figure*}[htbp]
\centering
\includegraphics[width=0.85\textwidth]{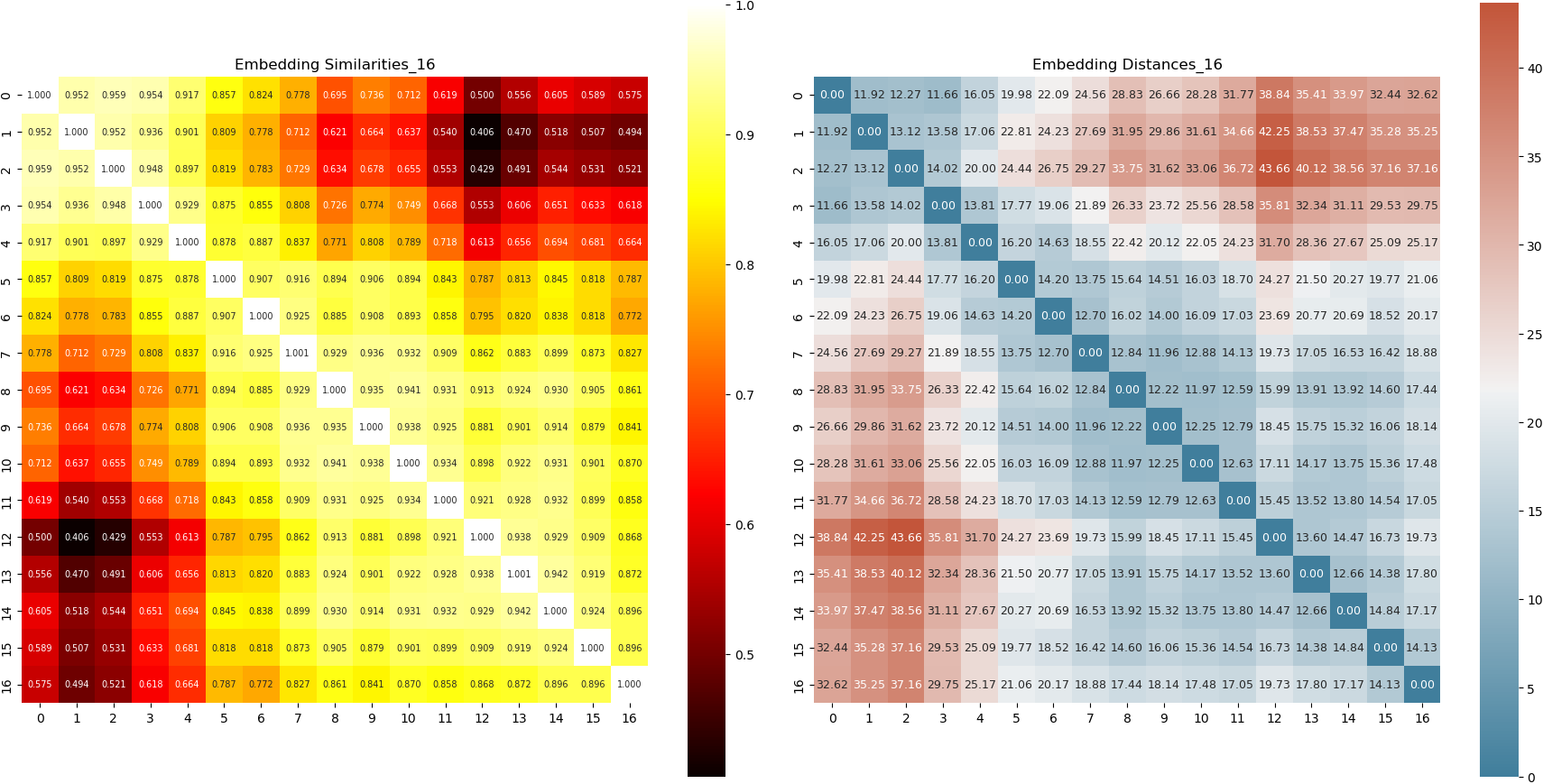}
\caption{12b Model, Step 16}
\label{fig:12b16}
\end{figure*}

\subsection{Supplementary Data for Embedding Dynamics}

\label{embedding dyanamics}
We have presented the PCA visualized embedding dynamics in Figure \ref{fig: embedding dyanamics}.
In this section, we provide the actual numbers of both Cosine Similarity and Euclidean distance to further illustrate this point.

Similarly to Figure \ref{fig: embedding dyanamics}, we also provide detailed cosine similarity and distance results from Figure \ref{fig:410m2} to Figure \ref{fig:12b16}.
From those figures, we can see that the cosine similarity fluctuates but remains relatively stable for sentences with different memorization scores.
Additionally, sentences that are not exactly the same but close regarding the memorization score are very close in the embedding space.
For example, fully memorized sentences are also close to sentences with high memorization scores and embedding similarity of over 0.9.
This shows the possibility that the model is generating paraphrased memorized sequences.
However, with the increase in the model size, the cosine similarity decreases.
For the Euclidean distance, we can see that the embedding distances have a decreasing trend with the increasing decoding steps, while the mutual Euclidean increases with the model size.
\begin{figure*}[htbp]
\centering
\includegraphics[width=\textwidth]{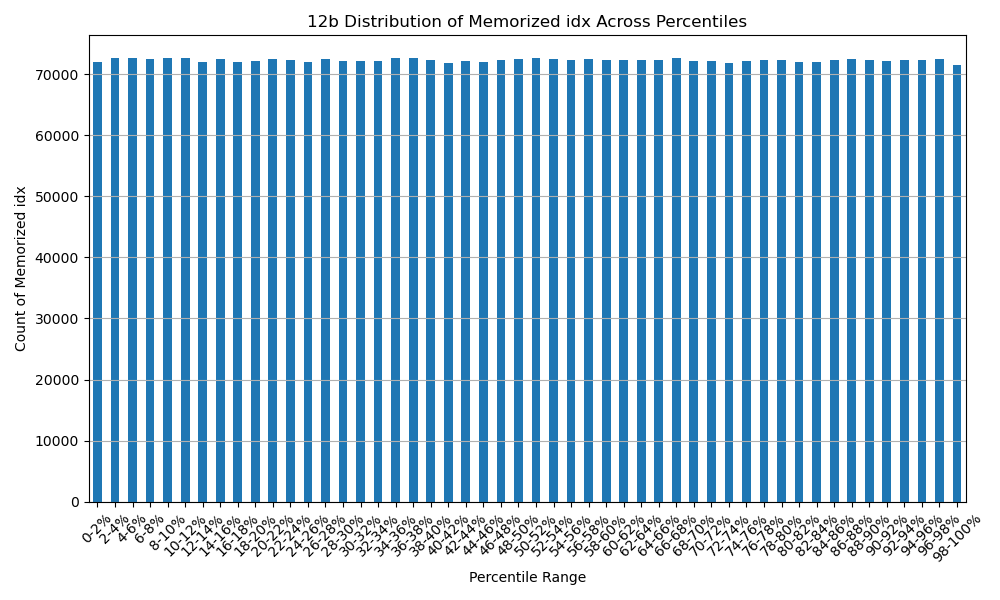}
\caption{Index Distribution of Memorized Sequences in 12b model.}
\label{fig:memorized index}
\end{figure*}
\subsection{Does LLM prefer to memorize specific parts within the training data?}
In this section, we discuss the question of whether LLM prefers to memorize a specific part within the training data.
We split the corpora into 50 parts based on their index and examine how many memorized sentences are in those parts.

From the result shown in Figure \ref{fig:memorized index}, though the number of memorized sentences is not completely evenly distributed, we can see that there is no significant part that the number of memorized sentences is clearly more than others.
This shows the training order does not affect memorization.

\subsection{Detail Gram Statistics}
We also provided detailed 1-gram, 2-gran, and 3-gram statistics for the memorized, un-memorized, half-memorized, and quarter-memorized contents. 
The average frequency is calculated at each step, and we compute the average frequency of the context, the average frequency of the continuation, and the average frequency of the whole sequence.
The results span from 70m size model to 12b size model shown in Table \ref{tab:1 gram 1} to Table \ref{tab:2 gram 2}.
\begin{table*}\small
\resizebox{\textwidth}{!}{%
\begin{tabular}{@{}lllllllllllllll@{}}
\toprule
target & memorized & memorized & memorized & memorized & memorized & memorized & memorized & forgotten & forgotten & forgotten & forgotten & forgotten & forgotten & forgotten \\
size & 70m & 160m & 410m & 1b & 2.8b & 6.9b & 12b & 70m & 160m & 410m & 1b & 2.8b & 6.9b & 12b \\
\bottomrule
0 & 1,761,877,325 & 1,744,677,836 & 1,753,317,716 & 1,749,565,941 & 1,759,384,624 & 1,763,400,000 & 1,794,900,000 & 1,732,531,426 & 1,732,526,814 & 1,733,883,954 & 1,735,430,248 & 1,757,800,000 & 1,717,300,000 & 1,753,200,000 \\
1 & 1,756,407,896 & 1,733,466,452 & 1,737,641,504 & 1,739,836,645 & 1,747,576,577 & 1,712,000,000 & 1,742,500,000 & 1,733,790,778 & 1,734,398,485 & 1,735,811,740 & 1,737,110,533 & 1,754,500,000 & 1,724,000,000 & 1,759,900,000 \\
2 & 1,754,544,442 & 1,735,828,900 & 1,739,272,135 & 1,740,794,079 & 1,750,863,001 & 1,758,000,000 & 1,804,000,000 & 1,733,955,892 & 1,734,433,227 & 1,735,927,786 & 1,737,263,808 & 1,719,800,000 & 1,730,300,000 & 1,755,600,000 \\
3 & 1,745,749,714 & 1,740,145,636 & 1,739,384,453 & 1,740,237,878 & 1,749,816,871 & 1,700,300,000 & 1,818,500,000 & 1,734,815,677 & 1,735,031,658 & 1,736,843,088 & 1,737,987,235 & 1,686,300,000 & 1,765,900,000 & 1,779,800,000 \\
4 & 1,733,568,727 & 1,729,484,889 & 1,728,938,412 & 1,733,592,149 & 1,742,113,248 & 1,766,400,000 & 1,708,800,000 & 1,735,128,356 & 1,735,452,086 & 1,737,120,799 & 1,738,793,702 & 1,767,300,000 & 1,733,300,000 & 1,759,200,000 \\
5 & 1,727,964,181 & 1,723,277,679 & 1,725,427,437 & 1,727,321,616 & 1,736,693,521 & 1,738,500,000 & 1,710,400,000 & 1,736,270,145 & 1,736,331,860 & 1,737,850,119 & 1,740,340,490 & 1,758,200,000 & 1,717,900,000 & 1,739,900,000 \\
6 & 1,725,841,351 & 1,716,563,761 & 1,723,081,626 & 1,725,702,469 & 1,735,415,386 & 1,758,000,000 & 1,742,900,000 & 1,735,559,708 & 1,736,435,200 & 1,738,820,970 & 1,740,165,047 & 1,780,700,000 & 1,742,500,000 & 1,713,400,000 \\
7 & 1,722,343,456 & 1,711,523,969 & 1,719,492,821 & 1,724,003,560 & 1,732,499,267 & 1,698,400,000 & 1,731,400,000 & 1,736,139,862 & 1,737,411,285 & 1,738,841,087 & 1,740,979,176 & 1,708,700,000 & 1,726,500,000 & 1,761,400,000 \\
8 & 1,715,952,971 & 1,703,541,696 & 1,712,534,695 & 1,717,230,465 & 1,730,417,914 & 1,679,100,000 & 1,709,000,000 & 1,736,938,342 & 1,738,000,206 & 1,739,984,548 & 1,741,715,014 & 1,752,400,000 & 1,775,600,000 & 1,740,500,000 \\
9 & 1,701,602,676 & 1,696,501,612 & 1,705,597,804 & 1,710,036,650 & 1,721,980,996 & 1,776,400,000 & 1,716,900,000 & 1,737,218,230 & 1,738,548,131 & 1,740,764,383 & 1,742,163,420 & 1,759,400,000 & 1,722,800,000 & 1,764,800,000 \\
10 & 1,708,709,485 & 1,703,507,469 & 1,707,974,902 & 1,712,561,959 & 1,724,060,121 & 1,672,300,000 & 1,733,400,000 & 1,737,059,426 & 1,738,812,313 & 1,740,973,658 & 1,742,911,460 & 1,760,200,000 & 1,807,900,000 & 1,753,800,000 \\
11 & 1,708,002,048 & 1,700,334,479 & 1,705,745,789 & 1,712,145,483 & 1,721,887,809 & 1,719,300,000 & 1,729,500,000 & 1,737,667,494 & 1,739,075,343 & 1,741,033,716 & 1,742,470,266 & 1,743,600,000 & 1,812,600,000 & 1,758,700,000 \\
12 & 1,704,765,267 & 1,690,528,200 & 1,700,680,032 & 1,703,727,332 & 1,714,582,320 & 1,739,700,000 & 1,711,800,000 & 1,737,833,800 & 1,738,990,635 & 1,741,004,691 & 1,742,710,814 & 1,768,300,000 & 1,754,700,000 & 1,771,800,000 \\
13 & 1,693,927,960 & 1,683,230,446 & 1,695,709,392 & 1,702,676,948 & 1,712,420,697 & 1,731,000,000 & 1,690,500,000 & 1,737,172,555 & 1,738,180,555 & 1,740,335,149 & 1,741,471,936 & 1,698,100,000 & 1,745,800,000 & 1,745,100,000 \\
14 & 1,696,634,570 & 1,681,852,317 & 1,695,295,636 & 1,701,256,215 & 1,712,257,191 & 1,725,700,000 & 1,699,700,000 & 1,737,522,088 & 1,738,958,841 & 1,740,701,751 & 1,742,059,832 & 1,740,500,000 & 1,699,000,000 & 1,728,500,000 \\
15 & 1,679,551,926 & 1,671,142,177 & 1,688,416,481 & 1,697,105,648 & 1,709,669,411 & 1,752,700,000 & 1,720,700,000 & 1,736,261,928 & 1,737,987,431 & 1,740,131,961 & 1,741,684,474 & 1,752,300,000 & 1,705,000,000 & 1,746,200,000 \\
16 & 1,667,903,751 & 1,670,413,255 & 1,678,747,207 & 1,691,187,011 & 1,704,670,219 & 1,719,900,000 & 1,699,800,000 & 1,735,533,670 & 1,737,210,612 & 1,738,579,355 & 1,740,344,603 & 1,750,300,000 & 1,781,300,000 & 1,752,600,000 \\
17 & 1,682,336,989 & 1,679,636,457 & 1,684,708,172 & 1,696,051,548 & 1,706,414,493 & 1,718,300,000 & 1,727,400,000 & 1,736,647,826 & 1,737,903,626 & 1,739,991,371 & 1,741,869,751 & 1,754,200,000 & 1,752,800,000 & 1,780,200,000 \\
18 & 1,696,024,732 & 1,688,157,613 & 1,693,550,892 & 1,701,124,435 & 1,711,161,137 & 1,689,300,000 & 1,685,300,000 & 1,736,875,723 & 1,738,169,859 & 1,740,186,111 & 1,742,150,532 & 1,723,700,000 & 1,764,500,000 & 1,720,000,000 \\
19 & 1,700,044,430 & 1,691,379,251 & 1,700,787,282 & 1,703,484,170 & 1,713,713,338 & 1,735,600,000 & 1,698,900,000 & 1,736,115,268 & 1,738,516,054 & 1,740,906,141 & 1,742,896,568 & 1,716,700,000 & 1,778,500,000 & 1,739,500,000 \\
20 & 1,710,839,951 & 1,700,701,910 & 1,706,890,340 & 1,710,112,768 & 1,717,662,092 & 1,703,300,000 & 1,750,400,000 & 1,735,637,929 & 1,737,519,099 & 1,740,228,710 & 1,741,971,117 & 1,776,900,000 & 1,761,300,000 & 1,752,300,000 \\
21 & 1,710,751,717 & 1,696,906,569 & 1,701,929,069 & 1,706,919,694 & 1,714,832,769 & 1,726,500,000 & 1,722,600,000 & 1,736,494,732 & 1,739,238,409 & 1,742,035,789 & 1,744,490,773 & 1,747,300,000 & 1,730,200,000 & 1,727,900,000 \\
22 & 1,707,031,048 & 1,694,596,312 & 1,700,606,809 & 1,707,393,546 & 1,714,688,122 & 1,688,200,000 & 1,719,400,000 & 1,735,343,187 & 1,738,714,382 & 1,741,858,543 & 1,743,975,469 & 1,763,800,000 & 1,738,800,000 & 1,757,900,000 \\
23 & 1,720,840,503 & 1,710,023,803 & 1,709,653,995 & 1,713,995,947 & 1,719,166,026 & 1,728,600,000 & 1,726,500,000 & 1,735,401,486 & 1,738,931,803 & 1,743,158,826 & 1,745,299,877 & 1,733,700,000 & 1,745,300,000 & 1,764,500,000 \\
24 & 1,731,576,174 & 1,715,940,520 & 1,714,003,336 & 1,715,023,292 & 1,717,926,947 & 1,758,400,000 & 1,722,100,000 & 1,734,596,940 & 1,738,386,519 & 1,742,505,032 & 1,744,500,841 & 1,749,300,000 & 1,761,700,000 & 1,760,400,000 \\
25 & 1,752,003,292 & 1,731,855,788 & 1,727,114,867 & 1,724,798,309 & 1,728,919,793 & 1,724,100,000 & 1,708,500,000 & 1,737,772,973 & 1,742,462,185 & 1,746,923,721 & 1,749,250,474 & 1,773,200,000 & 1,764,900,000 & 1,739,600,000 \\
26 & 1,755,434,381 & 1,735,918,727 & 1,728,805,371 & 1,726,260,802 & 1,725,855,889 & 1,717,700,000 & 1,710,400,000 & 1,736,254,689 & 1,741,221,711 & 1,746,794,790 & 1,750,257,624 & 1,782,600,000 & 1,737,100,000 & 1,737,500,000 \\
27 & 1,751,078,472 & 1,732,474,044 & 1,721,053,568 & 1,718,911,936 & 1,720,139,844 & 1,730,200,000 & 1,697,500,000 & 1,737,465,109 & 1,745,018,430 & 1,752,091,624 & 1,756,426,168 & 1,732,400,000 & 1,741,600,000 & 1,753,500,000 \\
28 & 1,752,459,582 & 1,738,590,437 & 1,733,534,889 & 1,725,443,806 & 1,726,659,853 & 1,686,300,000 & 1,679,200,000 & 1,727,033,252 & 1,738,018,143 & 1,747,326,896 & 1,754,100,491 & 1,764,800,000 & 1,752,900,000 & 1,767,200,000 \\
29 & 1,727,987,920 & 1,710,624,126 & 1,706,925,069 & 1,698,887,785 & 1,701,059,926 & 1,699,000,000 & 1,699,700,000 & 1,742,590,726 & 1,758,725,331 & 1,772,881,062 & 1,782,612,036 & 1,789,600,000 & 1,790,700,000 & 1,810,600,000 \\
30 & 1,723,009,291 & 1,712,857,933 & 1,705,443,834 & 1,699,764,161 & 1,701,619,618 & 1,711,900,000 & 1,710,500,000 & 1,742,138,296 & 1,764,672,256 & 1,781,900,380 & 1,793,731,882 & 1,863,100,000 & 1,825,200,000 & 1,818,400,000 \\
31 & 1,697,576,775 & 1,687,242,321 & 1,674,099,012 & 1,668,454,050 & 1,668,926,625 & 1,649,200,000 & 1,683,400,000 & 1,856,009,443 & 1,902,117,823 & 1,941,790,973 & 1,967,891,756 & 1,975,400,000 & 1,987,200,000 & 2,005,800,000 \\
Avg Context & 1,719,510,719 & 1,708,216,456 & 1,711,448,892 & 1,713,925,259 & 1,721,720,489 & 1,721,178,125 & 1,722,078,125 & 1,739,930,530 & 1,744,606,260 & 1,749,349,648 & 1,752,719,607 & 1,759,534,375 & 1,759,221,875 & 1,763,115,625 \\
32 & 1,812,724,891 & 1,801,299,369 & 1,783,560,981 & 1,771,127,395 & 1,765,849,310 & 1,739,200,000 & 1,722,000,000 & 957,760,109 & 963,964,253 & 981,205,278 & 986,644,973 & 1,017,100,000 & 1,024,300,000 & 989,330,000 \\
33 & 1,785,198,082 & 1,782,599,042 & 1,764,708,279 & 1,754,286,257 & 1,750,603,678 & 1,748,500,000 & 1,734,000,000 & 1,477,089,688 & 1,452,367,316 & 1,435,934,090 & 1,428,928,547 & 1,415,800,000 & 1,418,800,000 & 1,417,900,000 \\
34 & 1,772,561,111 & 1,767,491,083 & 1,753,649,952 & 1,744,671,079 & 1,742,356,129 & 1,769,300,000 & 1,735,800,000 & 1,623,949,166 & 1,609,532,509 & 1,597,061,854 & 1,592,797,469 & 1,560,600,000 & 1,576,000,000 & 1,592,600,000 \\
35 & 1,762,468,180 & 1,755,171,008 & 1,742,357,780 & 1,734,034,726 & 1,733,128,170 & 1,732,800,000 & 1,735,800,000 & 1,656,747,957 & 1,650,264,891 & 1,644,001,022 & 1,641,698,771 & 1,625,200,000 & 1,654,900,000 & 1,635,900,000 \\
36 & 1,751,084,744 & 1,744,819,021 & 1,731,757,083 & 1,727,988,589 & 1,726,739,343 & 1,728,600,000 & 1,706,600,000 & 1,678,449,622 & 1,673,912,651 & 1,669,885,767 & 1,668,314,472 & 1,642,100,000 & 1,683,100,000 & 1,643,100,000 \\
37 & 1,737,469,976 & 1,729,200,256 & 1,720,955,403 & 1,720,387,855 & 1,719,147,209 & 1,712,900,000 & 1,787,300,000 & 1,694,329,815 & 1,691,870,813 & 1,689,952,951 & 1,689,876,998 & 1,727,300,000 & 1,699,700,000 & 1,699,200,000 \\
38 & 1,737,353,894 & 1,724,369,474 & 1,716,962,208 & 1,710,620,923 & 1,711,850,544 & 1,691,900,000 & 1,761,400,000 & 1,696,281,632 & 1,695,900,348 & 1,694,668,623 & 1,695,468,826 & 1,674,600,000 & 1,697,300,000 & 1,679,200,000 \\
39 & 1,715,081,363 & 1,703,843,951 & 1,703,301,172 & 1,705,983,119 & 1,709,368,578 & 1,706,400,000 & 1,660,200,000 & 1,700,949,641 & 1,700,918,018 & 1,700,662,603 & 1,701,381,229 & 1,735,200,000 & 1,703,000,000 & 1,694,800,000 \\
40 & 1,686,685,834 & 1,682,653,925 & 1,690,744,526 & 1,694,008,976 & 1,698,953,947 & 1,698,400,000 & 1,732,100,000 & 1,699,950,424 & 1,700,320,136 & 1,700,698,424 & 1,701,888,972 & 1,709,600,000 & 1,749,200,000 & 1,717,200,000 \\
41 & 1,697,887,078 & 1,690,596,776 & 1,699,039,050 & 1,702,293,288 & 1,706,226,135 & 1,706,000,000 & 1,729,200,000 & 1,701,980,579 & 1,703,673,165 & 1,704,305,693 & 1,705,426,679 & 1,713,800,000 & 1,688,200,000 & 1,721,300,000 \\
42 & 1,706,446,759 & 1,699,625,915 & 1,704,257,959 & 1,710,138,829 & 1,711,973,193 & 1,706,800,000 & 1,696,400,000 & 1,699,916,580 & 1,701,963,362 & 1,704,179,446 & 1,704,239,602 & 1,693,500,000 & 1,680,700,000 & 1,699,700,000 \\
43 & 1,724,996,796 & 1,711,048,851 & 1,718,405,094 & 1,721,177,377 & 1,728,269,651 & 1,730,100,000 & 1,737,300,000 & 1,701,424,868 & 1,702,896,473 & 1,704,910,899 & 1,706,115,083 & 1,699,300,000 & 1,718,400,000 & 1,702,000,000 \\
44 & 1,729,978,559 & 1,717,877,194 & 1,721,748,175 & 1,727,317,540 & 1,725,798,564 & 1,752,400,000 & 1,680,700,000 & 1,699,235,317 & 1,702,231,574 & 1,704,286,186 & 1,705,328,377 & 1,732,700,000 & 1,719,100,000 & 1,715,900,000 \\
45 & 1,756,734,060 & 1,740,464,386 & 1,745,501,552 & 1,750,234,915 & 1,752,814,925 & 1,722,000,000 & 1,740,000,000 & 1,699,769,885 & 1,702,350,463 & 1,703,950,116 & 1,705,365,109 & 1,683,000,000 & 1,753,700,000 & 1,696,100,000 \\
46 & 1,772,357,723 & 1,765,976,491 & 1,769,251,508 & 1,775,699,434 & 1,780,492,152 & 1,796,400,000 & 1,773,900,000 & 1,697,965,233 & 1,700,236,541 & 1,702,146,238 & 1,703,508,158 & 1,740,600,000 & 1,727,900,000 & 1,724,700,000 \\
47 & 1,826,784,956 & 1,810,274,933 & 1,823,557,511 & 1,829,605,754 & 1,834,611,860 & 1,849,800,000 & 1,853,200,000 & 1,694,816,806 & 1,697,715,045 & 1,699,839,943 & 1,701,149,553 & 1,730,400,000 & 1,716,300,000 & 1,675,600,000 \\
Avg Continuation & 1,748,488,375 & 1,739,206,980 & 1,736,859,890 & 1,736,223,503 & 1,737,386,462 & 1,736,968,750 & 1,736,618,750 & 1,630,038,582 & 1,628,132,347 & 1,627,355,571 & 1,627,383,301 & 1,631,300,000 & 1,638,162,500 & 1,625,283,125 \\
Avg Sentence & 1,729,169,938 & 1,723,124,945 & 1,719,919,225 & 1,721,358,007 & 1,726,942,480 & 1,726,441,667 & 1,726,925,000 & 1,703,299,881 & 1,705,781,622 & 1,708,684,955 & 1,710,940,838 & 1,716,789,583 & 1,718,868,750 & 1,717,171,458 \\
\bottomrule
\end{tabular}%
}
\caption{One Gram Statistics for Memorized and Unmemorized Content}
\label{tab:1 gram 1}
\end{table*}

\begin{table*}\small
\resizebox{\textwidth}{!}{%
\begin{tabular}{@{}lllllllllllllll@{}}
\toprule
target & half & half & half & half & half & half & half & quarter & quarter & quarter & quarter & quarter & quarter & quarter \\
size & 70m & 160m & 410m & 1b & 2.8b & 6.9b & 12b & 70m & 160m & 410m & 1b & 2.8b & 6.9b & 12b \\
\bottomrule
0 & 1,747,500,000 & 1,780,200,000 & 1,752,900,000 & 1,700,100,000 & 1,729,500,000 & 1,733,700,000 & 1,715,500,000 & 1774800000 & 1,754,800,000 & 1785800000 & 1763800000 & 1781400000 & 1778900000 & 1756200000 \\
1 & 1,771,400,000 & 1,735,900,000 & 1,707,600,000 & 1,717,800,000 & 1,756,600,000 & 1,731,400,000 & 1,729,600,000 & 1746500000 & 1,781,100,000 & 1767400000 & 1746500000 & 1742600000 & 1779500000 & 1736400000 \\
2 & 1,747,800,000 & 1,745,400,000 & 1,758,500,000 & 1,731,800,000 & 1,705,000,000 & 1,687,200,000 & 1,774,100,000 & 1754000000 & 1,794,900,000 & 1781500000 & 1756500000 & 1768900000 & 1760300000 & 1753800000 \\
3 & 1,760,400,000 & 1,712,100,000 & 1,743,700,000 & 1,745,200,000 & 1,726,600,000 & 1,753,500,000 & 1,763,800,000 & 1791200000 & 1,776,600,000 & 1769500000 & 1776000000 & 1791900000 & 1748200000 & 1782600000 \\
4 & 1,693,800,000 & 1,720,700,000 & 1,726,500,000 & 1,696,900,000 & 1,748,200,000 & 1,697,300,000 & 1,723,500,000 & 1696800000 & 1,740,300,000 & 1789400000 & 1761900000 & 1792900000 & 1760800000 & 1754400000 \\
5 & 1,711,200,000 & 1,746,400,000 & 1,762,300,000 & 1,722,900,000 & 1,680,700,000 & 1,715,700,000 & 1,714,000,000 & 1761700000 & 1,765,300,000 & 1777000000 & 1720700000 & 1763800000 & 1765800000 & 1746100000 \\
6 & 1,700,900,000 & 1,717,100,000 & 1,673,600,000 & 1,753,500,000 & 1,774,500,000 & 1,735,800,000 & 1,771,000,000 & 1764900000 & 1,757,900,000 & 1726200000 & 1744400000 & 1742400000 & 1761400000 & 1785200000 \\
7 & 1,737,400,000 & 1,743,900,000 & 1,710,200,000 & 1,741,300,000 & 1,766,200,000 & 1,688,400,000 & 1,730,600,000 & 1755400000 & 1,690,800,000 & 1790000000 & 1717000000 & 1751600000 & 1740100000 & 1754900000 \\
8 & 1,705,800,000 & 1,721,300,000 & 1,728,200,000 & 1,723,200,000 & 1,720,900,000 & 1,676,100,000 & 1,738,200,000 & 1739900000 & 1,726,100,000 & 1799800000 & 1773900000 & 1806700000 & 1704900000 & 1737600000 \\
9 & 1,754,600,000 & 1,756,100,000 & 1,762,300,000 & 1,722,400,000 & 1,767,400,000 & 1,772,800,000 & 1,708,800,000 & 1785800000 & 1,741,600,000 & 1756200000 & 1763000000 & 1744900000 & 1757800000 & 1708900000 \\
10 & 1,737,700,000 & 1,726,600,000 & 1,761,000,000 & 1,715,500,000 & 1,724,600,000 & 1,754,200,000 & 1,757,500,000 & 1753600000 & 1,763,700,000 & 1757000000 & 1760600000 & 1748400000 & 1736600000 & 1770700000 \\
11 & 1,778,500,000 & 1,738,600,000 & 1,691,800,000 & 1,753,500,000 & 1,704,100,000 & 1,744,300,000 & 1,765,900,000 & 1774300000 & 1,773,600,000 & 1764500000 & 1746700000 & 1752300000 & 1741700000 & 1740200000 \\
12 & 1,721,500,000 & 1,696,300,000 & 1,678,400,000 & 1,699,000,000 & 1,714,100,000 & 1,718,900,000 & 1,765,400,000 & 1776700000 & 1,734,000,000 & 1771900000 & 1716500000 & 1736100000 & 1727900000 & 1730800000 \\
13 & 1,751,300,000 & 1,732,400,000 & 1,708,000,000 & 1,700,400,000 & 1,698,800,000 & 1,754,600,000 & 1,719,200,000 & 1714300000 & 1,744,900,000 & 1749500000 & 1741100000 & 1729200000 & 1746400000 & 1742400000 \\
14 & 1,737,600,000 & 1,731,900,000 & 1,735,500,000 & 1,682,700,000 & 1,695,200,000 & 1,701,600,000 & 1,730,800,000 & 1751000000 & 1,749,400,000 & 1731600000 & 1773600000 & 1738200000 & 1766800000 & 1712100000 \\
15 & 1,693,700,000 & 1,725,700,000 & 1,774,900,000 & 1,732,300,000 & 1,713,300,000 & 1,775,300,000 & 1,743,900,000 & 1728000000 & 1,724,300,000 & 1739000000 & 1742500000 & 1762500000 & 1736200000 & 1782100000 \\
16 & 1,739,000,000 & 1,749,900,000 & 1,740,800,000 & 1,777,700,000 & 1,747,300,000 & 1,749,300,000 & 1,702,800,000 & 1738500000 & 1,740,800,000 & 1705500000 & 1771500000 & 1729500000 & 1763500000 & 1749200000 \\
17 & 1,741,900,000 & 1,694,200,000 & 1,742,100,000 & 1,771,200,000 & 1,721,200,000 & 1,741,300,000 & 1,783,000,000 & 1745700000 & 1,740,100,000 & 1767500000 & 1736700000 & 1738900000 & 1741000000 & 1761400000 \\
18 & 1,738,200,000 & 1,703,300,000 & 1,740,700,000 & 1,711,900,000 & 1,705,700,000 & 1,704,600,000 & 1,686,400,000 & 1737100000 & 1,743,400,000 & 1743200000 & 1737100000 & 1724500000 & 1735100000 & 1756800000 \\
19 & 1,728,500,000 & 1,725,600,000 & 1,729,100,000 & 1,719,600,000 & 1,725,700,000 & 1,679,600,000 & 1,747,000,000 & 1726900000 & 1,791,300,000 & 1728300000 & 1740300000 & 1707300000 & 1736600000 & 1719900000 \\
20 & 1,691,300,000 & 1,761,400,000 & 1,714,800,000 & 1,761,200,000 & 1,695,900,000 & 1,702,800,000 & 1,733,400,000 & 1727700000 & 1,767,400,000 & 1707000000 & 1730800000 & 1771400000 & 1725200000 & 1795800000 \\
21 & 1,677,600,000 & 1,738,600,000 & 1,699,400,000 & 1,737,100,000 & 1,691,400,000 & 1,680,900,000 & 1,697,700,000 & 1755200000 & 1,741,500,000 & 1750000000 & 1777700000 & 1751000000 & 1741700000 & 1746900000 \\
22 & 1,740,300,000 & 1,713,600,000 & 1,716,200,000 & 1,668,000,000 & 1,698,100,000 & 1,718,800,000 & 1,722,600,000 & 1774400000 & 1,710,600,000 & 1757000000 & 1774700000 & 1797700000 & 1752900000 & 1746100000 \\
23 & 1,748,200,000 & 1,666,100,000 & 1,664,400,000 & 1,729,100,000 & 1,677,100,000 & 1,734,500,000 & 1,731,200,000 & 1747300000 & 1,737,000,000 & 1766600000 & 1782700000 & 1761900000 & 1757800000 & 1723400000 \\
24 & 1,691,700,000 & 1,735,300,000 & 1,716,500,000 & 1,722,300,000 & 1,733,100,000 & 1,699,600,000 & 1,678,600,000 & 1735800000 & 1,734,200,000 & 1748600000 & 1712500000 & 1743000000 & 1772400000 & 1700900000 \\
25 & 1,728,800,000 & 1,656,600,000 & 1,726,600,000 & 1,700,200,000 & 1,690,500,000 & 1,697,900,000 & 1,732,800,000 & 1731200000 & 1,766,800,000 & 1740300000 & 1724100000 & 1736800000 & 1782600000 & 1718200000 \\
26 & 1,700,700,000 & 1,715,200,000 & 1,654,200,000 & 1,694,200,000 & 1,703,700,000 & 1,684,400,000 & 1,733,500,000 & 1708800000 & 1,715,100,000 & 1739000000 & 1734800000 & 1721100000 & 1695100000 & 1749000000 \\
27 & 1,724,500,000 & 1,645,400,000 & 1,754,400,000 & 1,707,800,000 & 1,724,900,000 & 1,721,800,000 & 1,672,900,000 & 1703300000 & 1,715,900,000 & 1711100000 & 1697200000 & 1718800000 & 1708100000 & 1695900000 \\
28 & 1,658,300,000 & 1,712,500,000 & 1,744,100,000 & 1,660,800,000 & 1,739,500,000 & 1,678,300,000 & 1,701,000,000 & 1743400000 & 1,732,800,000 & 1752700000 & 1712300000 & 1724300000 & 1736400000 & 1680500000 \\
29 & 1,692,600,000 & 1,667,500,000 & 1,686,700,000 & 1,655,700,000 & 1,657,100,000 & 1,676,700,000 & 1,707,500,000 & 1697300000 & 1,771,300,000 & 1731300000 & 1717800000 & 1677500000 & 1714100000 & 1691500000 \\
30 & 1,655,500,000 & 1,690,700,000 & 1,658,200,000 & 1,647,300,000 & 1,683,200,000 & 1,639,600,000 & 1,628,200,000 & 1760800000 & 1,689,800,000 & 1675800000 & 1691500000 & 1678700000 & 1649800000 & 1697800000 \\
31 & 1,666,800,000 & 1,541,100,000 & 1,637,100,000 & 1,564,000,000 & 1,513,400,000 & 1,579,400,000 & 1,554,100,000 & 1562100000 & 1,515,300,000 & 1536200000 & 1484100000 & 1477500000 & 1436800000 & 1507600000 \\
Avg Context & 1,721,093,750 & 1,713,987,500 & 1,718,771,875 & 1,711,456,250 & 1,710,421,875 & 1,710,321,875 & 1,720,765,625 & 1,739,512,500 & 1,738,518,750 & 1,744,262,500 & 1,735,328,125 & 1,737,928,125 & 1,733,200,000 & 1,732,353,125 \\
32 & 1,878,300,000 & 1,871,300,000 & 1,800,300,000 & 1,834,700,000 & 1,812,500,000 & 1,719,500,000 & 1,791,700,000 & 2189100000 & 2,090,900,000 & 2020500000 & 1992900000 & 1907700000 & 1938700000 & 1895000000 \\
33 & 1,842,000,000 & 1,854,700,000 & 1,836,200,000 & 1,829,200,000 & 1,869,400,000 & 1,834,500,000 & 1,749,400,000 & 2458900000 & 2,344,000,000 & 2383700000 & 2295800000 & 2224900000 & 2269700000 & 2275200000 \\
34 & 1,842,600,000 & 1,900,800,000 & 1,843,500,000 & 1,862,600,000 & 1,778,400,000 & 1,812,100,000 & 1,832,600,000 & 2650800000 & 2,597,500,000 & 2592700000 & 2537800000 & 2483300000 & 2511000000 & 2568800000 \\
35 & 1,841,600,000 & 1,875,500,000 & 1,883,700,000 & 1,894,300,000 & 1,866,800,000 & 1,843,400,000 & 1,868,600,000 & 2520900000 & 2,583,300,000 & 2586100000 & 2591600000 & 2586700000 & 2599000000 & 2572600000 \\
36 & 1,871,800,000 & 1,901,500,000 & 1,932,600,000 & 1,931,300,000 & 1,926,100,000 & 1,872,300,000 & 1,893,200,000 & 1778900000 & 1,742,400,000 & 1743700000 & 1696700000 & 1687300000 & 1715400000 & 1720500000 \\
37 & 1,885,600,000 & 1,921,800,000 & 1,997,900,000 & 1,915,800,000 & 1,981,000,000 & 2,077,100,000 & 2,038,100,000 & 1736300000 & 1,739,900,000 & 1728100000 & 1685100000 & 1664500000 & 1681500000 & 1700300000 \\
38 & 1,998,200,000 & 2,101,300,000 & 2,081,700,000 & 2,115,900,000 & 2,156,700,000 & 2,209,100,000 & 2,254,500,000 & 1762300000 & 1,710,100,000 & 1711100000 & 1737600000 & 1718500000 & 1736000000 & 1705300000 \\
39 & 2,032,100,000 & 2,158,400,000 & 2,168,100,000 & 2,189,200,000 & 2,246,300,000 & 2,276,200,000 & 2,369,000,000 & 1809200000 & 1,768,100,000 & 1799700000 & 1747600000 & 1744100000 & 1738300000 & 1758800000 \\
40 & 1,723,200,000 & 1,690,600,000 & 1,715,300,000 & 1,692,500,000 & 1,691,600,000 & 1,691,900,000 & 1,719,100,000 & 1765200000 & 1,751,200,000 & 1789500000 & 1746300000 & 1757000000 & 1733900000 & 1779400000 \\
41 & 1,738,700,000 & 1,731,600,000 & 1,699,700,000 & 1,709,900,000 & 1,706,600,000 & 1,668,700,000 & 1,706,000,000 & 1773400000 & 1,777,100,000 & 1770200000 & 1803300000 & 1780300000 & 1788700000 & 1780500000 \\
42 & 1,707,400,000 & 1,795,200,000 & 1,719,700,000 & 1,721,300,000 & 1,743,300,000 & 1,700,900,000 & 1,725,800,000 & 1822600000 & 1,814,600,000 & 1763300000 & 1800800000 & 1793500000 & 1788700000 & 1740000000 \\
43 & 1751700000 & 1,708,000,000 & 1,720,500,000 & 1,713,600,000 & 1,733,400,000 & 1,727,600,000 & 1,687,200,000 & 1839800000 & 1,774,600,000 & 1838300000 & 1761600000 & 1754700000 & 1787300000 & 1805000000 \\
44 & 1753600000 & 1,742,000,000 & 1,764,800,000 & 1,694,000,000 & 1,739,700,000 & 1,686,500,000 & 1,721,000,000 & 1760300000 & 1,806,500,000 & 1817700000 & 1760600000 & 1787800000 & 1782700000 & 1752900000 \\
45 & 1729900000 & 1,720,800,000 & 1,715,200,000 & 1,721,700,000 & 1,786,500,000 & 1,697,200,000 & 1,733,700,000 & 1775700000 & 1,763,900,000 & 1829800000 & 1840800000 & 1784300000 & 1822300000 & 1734000000 \\
46 & 1733000000 & 1,717,200,000 & 1,741,500,000 & 1,705,100,000 & 1,750,600,000 & 1,721,000,000 & 1,737,400,000 & 1797500000 & 1,799,100,000 & 1788600000 & 1729000000 & 1772000000 & 1765100000 & 1758700000 \\
47 & 1732500000 & 1,715,600,000 & 1,748,500,000 & 1,788,200,000 & 1,710,100,000 & 1,731,500,000 & 1,712,500,000 & 1763400000 & 1,757,300,000 & 1783200000 & 1835300000 & 1791700000 & 1764800000 & 1766000000 \\
Avg Continuation & 1,816,387,500 & 1,837,893,750 & 1,835,575,000 & 1,832,456,250 & 1,843,687,500 & 1,829,343,750 & 1,846,237,500 & 1,950,268,750 & 1,926,281,250 & 1,934,137,500 & 1,910,175,000 & 1,889,893,750 & 1,901,443,750 & 1,894,562,500 \\
Avg Sentence & 1,752,858,333 & 1,755,289,583 & 1,757,706,250 & 1,751,789,583 & 1,754,843,750 & 1,749,995,833 & 1,762,589,583 & 1,809,764,583 & 1,801,106,250 & 1,807,554,167 & 1,793,610,417 & 1,788,583,333 & 1,789,281,250 & 1,786,422,917 \\
\bottomrule
\end{tabular}%
}
\caption{One Gram Statistics for Half-memorized and Quarter-memorized content}
\label{tab:1 gram 2}
\end{table*}

\begin{table*}\small
\resizebox{\textwidth}{!}{%
\begin{tabular}{@{}llllllllllllllll@{}}
\toprule
target & memorized & memorized & memorized & memorized & memorized & memorized & memorized & forgotten & forgotten & forgotten & forgotten & forgotten & forgotten & forgotten \\ 
size & 70m & 160m & 410m & 1b & 2.8b & 6.9b & 12b & 70m & 160m & 410m & 1b & 2.8b & 6.9b & 12b \\
\midrule
1 & 56,865,298 & 55,633,408 & 57,631,394 & 57,851,391 & 58,747,419 & 59,390,492 & 59,538,566 & 67,723,858 & 67,753,352 & 67,683,530 & 67,708,452 & 67,740,430 & 67,754,040 & 67,767,823 \\
2 & 57,323,714 & 56,317,639 & 57,433,612 & 57,993,010 & 58,602,181 & 59,488,016 & 59,803,389 & 67,737,009 & 67,768,452 & 67,769,364 & 67,815,489 & 67,769,957 & 67,800,825 & 67,828,380 \\
3 & 56,730,613 & 56,231,828 & 57,388,365 & 58,009,821 & 58,928,771 & 59,525,329 & 59,598,379 & 67,775,953 & 67,738,481 & 67,807,016 & 67,780,588 & 67,804,139 & 67,851,921 & 67,894,036 \\
4 & 55,911,242 & 56,234,359 & 56,980,241 & 57,792,288 & 58,806,766 & 59,391,815 & 59,636,502 & 67,949,220 & 67,825,207 & 67,839,517 & 67,841,697 & 67,867,488 & 67,909,524 & 67,967,557 \\
5 & 55,692,756 & 55,655,983 & 56,819,836 & 57,273,741 & 58,351,556 & 59,022,366 & 59,346,451 & 67,958,033 & 67,886,128 & 67,874,628 & 68,003,000 & 67,965,687 & 68,045,776 & 68,030,203 \\
6 & 55,894,920 & 55,666,111 & 56,911,193 & 57,438,723 & 58,686,676 & 59,320,405 & 59,557,076 & 67,985,473 & 67,869,913 & 67,917,993 & 68,012,109 & 67,974,605 & 68,003,875 & 68,007,874 \\
7 & 56,047,068 & 55,753,285 & 56,733,219 & 57,360,465 & 58,300,186 & 59,087,792 & 59,572,316 & 68,058,487 & 68,015,919 & 68,064,231 & 68,101,370 & 68,115,977 & 68,153,223 & 68,184,607 \\
8 & 55,832,397 & 55,349,442 & 56,329,311 & 57,111,974 & 58,349,971 & 58,924,645 & 59,505,541 & 68,017,225 & 68,013,215 & 68,034,561 & 68,111,358 & 68,101,007 & 68,139,029 & 68,201,839 \\
9 & 55,253,643 & 54,713,056 & 55,885,459 & 56,532,115 & 57,837,312 & 58,610,331 & 59,086,863 & 68,135,592 & 68,114,172 & 68,143,152 & 68,214,493 & 68,138,035 & 68,189,743 & 68,155,354 \\
10 & 54,503,716 & 54,257,056 & 55,495,149 & 56,166,844 & 57,307,569 & 58,022,654 & 58,377,791 & 68,160,293 & 68,172,545 & 68,187,816 & 68,250,398 & 68,215,884 & 68,267,972 & 68,242,452 \\
11 & 54,988,987 & 54,708,110 & 55,773,288 & 56,333,064 & 57,701,866 & 58,500,535 & 58,871,367 & 68,234,320 & 68,217,765 & 68,269,026 & 68,248,385 & 68,243,497 & 68,268,692 & 68,241,732 \\
12 & 54,935,679 & 54,664,019 & 55,533,124 & 56,267,162 & 57,079,217 & 58,003,846 & 58,331,501 & 68,213,578 & 68,234,560 & 68,252,214 & 68,247,968 & 68,328,780 & 68,343,161 & 68,299,170 \\
13 & 54,767,327 & 54,145,183 & 55,296,145 & 55,857,794 & 56,871,961 & 57,652,393 & 57,921,174 & 68,226,649 & 68,274,620 & 68,277,162 & 68,277,302 & 68,337,296 & 68,382,126 & 68,389,974 \\
14 & 54,021,320 & 53,569,285 & 54,624,126 & 55,455,249 & 56,409,277 & 57,387,470 & 57,829,909 & 68,213,386 & 68,233,133 & 68,245,257 & 68,296,277 & 68,307,432 & 68,429,171 & 68,395,506 \\
15 & 54,367,150 & 53,970,143 & 55,279,001 & 55,940,451 & 57,010,866 & 57,759,141 & 58,109,891 & 68,177,596 & 68,207,430 & 68,237,580 & 68,241,325 & 68,279,727 & 68,326,632 & 68,368,250 \\
16 & 53,451,225 & 53,327,809 & 54,393,699 & 55,350,104 & 56,364,900 & 57,290,723 & 57,711,596 & 68,022,941 & 68,094,829 & 68,169,848 & 68,158,831 & 68,224,797 & 68,273,585 & 68,255,991 \\
17 & 53,645,203 & 53,634,633 & 54,177,271 & 55,078,600 & 56,138,436 & 57,251,531 & 57,541,351 & 68,138,549 & 68,147,333 & 68,221,638 & 68,287,831 & 68,325,084 & 68,415,608 & 68,411,921 \\
18 & 54,668,363 & 54,174,847 & 55,256,176 & 56,064,089 & 56,740,259 & 57,769,268 & 57,973,134 & 68,224,597 & 68,282,517 & 68,395,059 & 68,437,841 & 68,522,898 & 68,598,086 & 68,599,844 \\
19 & 55,562,101 & 54,791,623 & 55,294,695 & 55,928,589 & 56,768,136 & 57,599,180 & 58,081,287 & 68,209,538 & 68,324,444 & 68,446,514 & 68,525,618 & 68,628,926 & 68,686,765 & 68,753,261 \\
20 & 55,938,657 & 55,081,208 & 55,877,542 & 56,253,266 & 57,048,367 & 57,967,318 & 58,106,207 & 68,136,189 & 68,313,557 & 68,470,777 & 68,543,198 & 68,639,292 & 68,722,909 & 68,829,286 \\
21 & 56,331,087 & 55,229,266 & 55,652,638 & 56,067,216 & 56,679,203 & 57,502,276 & 57,635,297 & 68,141,999 & 68,361,062 & 68,547,360 & 68,644,264 & 68,783,591 & 68,859,312 & 68,950,347 \\
22 & 56,997,804 & 55,715,430 & 56,053,895 & 56,356,954 & 56,761,453 & 57,545,255 & 57,751,704 & 68,198,529 & 68,480,921 & 68,670,594 & 68,818,205 & 69,044,454 & 69,163,424 & 69,194,391 \\
23 & 56,802,977 & 55,429,313 & 55,735,965 & 56,197,410 & 56,471,773 & 57,057,789 & 57,383,903 & 68,143,528 & 68,525,019 & 68,831,257 & 68,946,803 & 69,230,759 & 69,321,426 & 69,445,410 \\
24 & 57,826,178 & 56,183,113 & 56,044,365 & 55,945,228 & 55,984,317 & 56,618,627 & 56,771,473 & 68,271,377 & 68,728,185 & 69,037,189 & 69,197,403 & 69,488,552 & 69,616,627 & 69,731,068 \\
25 & 58,617,239 & 57,079,668 & 56,202,206 & 56,102,928 & 55,885,915 & 56,296,069 & 56,349,525 & 68,384,381 & 68,871,654 & 69,255,113 & 69,446,599 & 69,840,923 & 69,999,312 & 70,097,498 \\
26 & 59,309,065 & 57,207,648 & 56,075,430 & 55,816,944 & 55,552,723 & 55,814,962 & 55,846,995 & 68,608,814 & 69,166,616 & 69,653,434 & 69,885,998 & 70,350,797 & 70,565,524 & 70,654,203 \\
27 & 59,214,372 & 56,581,831 & 55,453,438 & 55,266,849 & 54,772,970 & 55,091,733 & 55,084,942 & 68,827,806 & 69,591,707 & 70,264,503 & 70,708,424 & 71,312,980 & 71,558,722 & 71,777,086 \\
28 & 58,905,505 & 56,371,887 & 55,174,878 & 54,389,172 & 54,041,511 & 54,335,228 & 54,268,336 & 68,505,025 & 69,619,124 & 70,499,723 & 71,115,782 & 71,913,592 & 72,172,923 & 72,454,327 \\
29 & 58,023,696 & 56,122,492 & 55,053,358 & 53,998,472 & 53,301,028 & 53,274,003 & 53,087,300 & 68,732,220 & 70,492,616 & 71,644,050 & 72,524,388 & 73,553,428 & 73,967,910 & 74,302,156 \\
30 & 56,113,860 & 54,313,900 & 52,636,358 & 51,531,025 & 50,770,809 & 50,519,079 & 50,314,047 & 74,810,707 & 77,539,372 & 79,590,655 & 80,895,889 & 82,592,009 & 83,417,356 & 83,993,892 \\
31 & 51,887,230 & 50,597,329 & 48,600,945 & 47,771,643 & 46,600,073 & 46,175,796 & 45,915,086 & 77,719,313 & 81,332,604 & 84,241,907 & 86,088,401 & 88,563,161 & 89,838,828 & 90,658,855\\
Avg Context & 56,013,884 & 55,119,707 & 55,541,817 & 55,854,922 & 56,415,273 & 57,038,583 & 57,255,126 & 68,698,135 & 69,103,111 & 69,436,860 & 69,657,280 & 69,942,103 & 70,098,195 & 70,196,268 \\
32 & 55,823,256 & 53,884,097 & 51,681,052 & 50,569,791 & 49,394,788 & 48,863,702 & 48,537,557 & 13,930,348 & 15,076,103 & 16,016,778 & 16,465,745 & 16,788,294 & 17,279,969 & 17,455,400 \\
33 & 58,738,344 & 56,674,056 & 54,054,542 & 52,888,037 & 51,528,984 & 50,887,306 & 50,414,778 & 26,816,835 & 27,093,410 & 27,773,603 & 27,962,715 & 28,191,993 & 28,477,352 & 28,654,271 \\
34 & 56,938,764 & 55,281,907 & 52,909,883 & 51,461,738 & 50,038,303 & 49,641,208 & 49,084,752 & 47,564,034 & 45,235,994 & 44,031,407 & 43,883,655 & 43,877,411 & 43,554,253 & 43,485,497 \\
35 & 54,366,396 & 53,029,014 & 50,888,507 & 49,755,655 & 48,770,433 & 48,334,318 & 47,981,902 & 53,101,160 & 51,120,417 & 49,742,735 & 49,185,432 & 48,518,180 & 48,081,379 & 47,894,499\\
36 & 52,936,332 & 51,682,636 & 49,570,706 & 48,713,389 & 47,794,113 & 47,573,073 & 47,233,332 & 58,003,949 & 56,939,073 & 55,963,819 & 55,597,113 & 55,051,241 & 54,774,326 & 54,588,795 \\
37 & 51,647,223 & 50,371,630 & 48,303,608 & 47,632,078 & 46,858,461 & 46,786,053 & 46,376,421 & 61,213,159 & 60,453,961 & 59,829,451 & 59,536,354 & 59,153,093 & 59,031,495 & 58,838,218 \\
38 & 50,931,788 & 49,209,356 & 47,576,761 & 46,731,794 & 45,968,874 & 45,790,581 & 45,477,362 & 62,655,390 & 62,137,314 & 61,620,255 & 61,458,679 & 61,231,410 & 61,110,159 & 61,003,516 \\
39 & 51,042,601 & 49,027,546 & 47,618,052 & 46,768,291 & 45,998,476 & 45,883,841 & 45,473,705 & 63,348,201 & 62,998,381 & 62,683,843 & 62,649,981 & 62,440,555 & 62,378,970 & 62,319,582 \\
40 & 49,590,360 & 47,756,219 & 46,431,515 & 45,965,206 & 45,449,164 & 45,252,564 & 45,184,300 & 63,866,857 & 63,600,670 & 63,423,587 & 63,430,422 & 63,301,531 & 63,282,617 & 63,207,398 \\
41 & 49,954,374 & 47,928,574 & 46,848,807 & 46,364,193 & 45,679,210 & 45,474,688 & 45,440,294 & 64,301,293 & 64,110,817 & 63,913,386 & 63,893,724 & 63,873,160 & 63,880,434 & 63,857,697 \\
42 & 50,897,285 & 48,822,840 & 47,723,273 & 47,389,348 & 46,461,956 & 46,336,541 & 46,106,581 & 64,486,391 & 64,515,591 & 64,453,381 & 64,379,111 & 64,406,903 & 64,442,057 & 64,385,303 \\
43 & 52,467,447 & 49,948,494 & 48,687,532 & 48,087,823 & 47,426,244 & 47,134,027 & 46,970,169 & 64,557,953 & 64,566,576 & 64,532,662 & 64,488,626 & 64,565,552 & 64,573,767 & 64,585,706  \\
44 & 54,876,822 & 52,145,505 & 50,513,749 & 49,816,893 & 48,819,838 & 48,428,789 & 48,281,333 & 64,662,055 & 64,739,419 & 64,747,685 & 64,807,289 & 64,861,456 & 64,982,075 & 64,897,827 \\
45 & 56,056,772 & 53,288,677 & 51,530,428 & 50,709,794 & 49,538,754 & 49,171,650 & 48,847,311 & 64,772,578 & 64,907,757 & 64,948,826 & 64,997,734 & 65,068,658 & 65,124,986 & 65,132,858  \\
46 & 61,167,369 & 59,469,476 & 57,204,341 & 55,850,490 & 54,489,360 & 53,898,637 & 53,288,932 & 64,727,227 & 64,840,307 & 64,926,333 & 64,908,159 & 65,053,638 & 65,091,642 & 65,154,625 \\
47 & 71,009,467 & 67,839,176 & 66,565,545 & 65,503,549 & 63,999,825 & 63,292,284 & 62,803,150 & 64,590,446 & 64,867,523 & 64,840,482 & 64,903,198 & 65,014,873 & 65,036,948 & 65,120,427 \\
48 & 66,672,308 & 64,419,151 & 63,096,912 & 62,676,497 & 61,761,157 & 60,991,890 & 60,750,602 & 66,612,523 & 66,803,144 & 66,905,353 & 66,952,648 & 67,075,374 & 67,116,770 & 67,266,114 \\
Avg Continuation & 55,595,112 & 53,575,197 & 51,835,601 & 50,993,210 & 49,998,702 & 49,631,832 & 49,308,969 & 57,012,377 & 56,706,262 & 56,491,387 & 56,441,211 & 56,380,784 & 56,365,835 & 56,343,984 \\
Avg Sentence & 55,865,569 & 54,572,693 & 54,229,199 & 54,133,066 & 54,142,738 & 54,415,359 & 54,440,862 & 64,559,429 & 64,712,561 & 64,852,005 & 64,976,589 & 65,139,135 & 65,234,651 & 65,290,251 \\ \bottomrule
\end{tabular}%
}
\caption{2 Gram Statistics for memorized and unmemorized content}
\label{tab:2 gram 1}
\end{table*}

\begin{table*}\small
\resizebox{\textwidth}{!}{%
\begin{tabular}{@{}lllllllllllllll@{}}
\toprule
target & half & half & half & half & half & half & half & quarter & quarter & quarter & quarter & quarter & quarter & quarter \\
size & 70m & 160m & 410m & 1b & 2.8b & 6.9b & 12b & 70m & 160m & 410m & 1b & 2.8b & 6.9b & 12b \\
\midrule
1 & 53,888,945 & 55,774,256 & 56,835,849 & 57,399,791 & 58,749,349 & 58,746,839 & 59,572,656 & 65,365,004 & 66,059,083 & 66,813,871 & 67,057,786 & 67,482,982 & 67,249,886 & 67,522,493 \\
2 & 53,633,778 & 55,585,291 & 56,294,740 & 57,125,864 & 58,375,426 & 58,787,083 & 59,024,756 & 65,505,804 & 66,197,266 & 66,585,597 & 66,774,723 & 67,277,478 & 67,415,528 & 67,313,805 \\
3 & 53,966,515 & 55,464,751 & 56,571,797 & 57,248,585 & 58,522,726 & 59,007,609 & 59,282,618 & 65,526,925 & 66,301,214 & 66,609,558 & 67,101,450 & 67,242,630 & 67,240,433 & 67,479,250 \\
4 & 53,788,979 & 55,012,249 & 55,852,524 & 56,732,428 & 58,632,577 & 58,585,481 & 59,106,526 & 65,575,121 & 65,977,326 & 66,498,055 & 66,914,635 & 67,144,082 & 67,128,936 & 67,304,810 \\
5 & 52,887,357 & 54,130,559 & 55,380,593 & 56,487,949 & 57,789,441 & 58,410,016 & 58,836,366 & 65,413,320 & 66,086,522 & 66,640,160 & 66,821,619 & 67,103,545 & 67,257,537 & 67,368,841 \\
6 & 53,346,585 & 54,626,718 & 56,348,320 & 56,603,397 & 58,159,466 & 58,648,425 & 58,945,615 & 65,113,237 & 66,153,641 & 66,499,304 & 66,682,733 & 67,063,613 & 67,252,679 & 67,384,155 \\
7 & 52,836,947 & 54,888,957 & 55,457,422 & 56,613,020 & 57,855,215 & 58,349,009 & 58,580,014 & 65,090,106 & 65,649,597 & 66,459,689 & 66,678,897 & 66,947,213 & 67,002,958 & 67,127,891 \\
8 & 52,478,474 & 54,201,487 & 55,628,345 & 56,114,532 & 57,493,118 & 58,018,440 & 58,920,964 & 64,824,999 & 65,498,431 & 66,241,656 & 66,399,994 & 66,746,465 & 66,760,013 & 66,819,417 \\
9 & 53,600,100 & 55,410,940 & 56,362,313 & 57,618,326 & 58,256,347 & 59,088,717 & 59,814,719 & 64,941,905 & 65,788,899 & 66,460,344 & 66,506,701 & 66,921,900 & 67,080,681 & 67,174,870 \\
10 & 53,868,058 & 55,087,199 & 56,362,786 & 57,058,982 & 58,314,177 & 58,745,840 & 59,526,404 & 64,563,380 & 65,502,285 & 66,335,635 & 66,276,258 & 66,961,511 & 66,922,326 & 66,965,502 \\
11 & 52,884,280 & 54,383,701 & 56,075,362 & 56,818,188 & 58,050,191 & 58,352,813 & 58,596,784 & 64,715,807 & 65,616,229 & 66,029,659 & 66,452,694 & 66,712,759 & 66,674,522 & 66,995,794 \\
12 & 52,928,646 & 54,849,387 & 55,982,049 & 56,771,448 & 58,094,228 & 58,565,819 & 58,811,948 & 64,820,938 & 65,424,995 & 65,790,449 & 66,253,435 & 66,525,526 & 66,698,298 & 66,719,967 \\
13 & 52,690,921 & 54,559,756 & 55,697,194 & 56,532,692 & 57,846,992 & 58,355,202 & 58,946,804 & 64,842,590 & 65,142,936 & 65,893,506 & 66,065,102 & 66,465,922 & 66,738,926 & 66,653,765 \\
14 & 52,781,679 & 54,368,904 & 56,019,791 & 56,847,208 & 57,869,958 & 58,242,828 & 58,685,929 & 64,404,765 & 65,245,584 & 65,797,393 & 66,052,000 & 66,479,573 & 66,541,779 & 66,586,451 \\
15 & 53,765,298 & 55,537,152 & 56,289,403 & 57,279,213 & 58,113,960 & 58,741,123 & 59,494,649 & 64,231,835 & 64,970,624 & 65,603,603 & 65,691,554 & 66,258,007 & 66,393,008 & 66,421,189 \\
16 & 53,740,078 & 55,037,415 & 55,929,818 & 56,941,982 & 57,702,245 & 58,282,107 & 58,697,163 & 64,285,467 & 64,973,050 & 65,644,470 & 65,985,554 & 66,366,943 & 66,449,639 & 66,680,453 \\
17 & 53,610,921 & 55,130,291 & 56,781,859 & 57,648,192 & 58,161,184 & 58,670,246 & 59,044,372 & 64,584,535 & 65,324,214 & 66,141,495 & 66,173,896 & 66,423,703 & 66,634,663 & 66,523,072 \\
18 & 52,403,931 & 54,507,436 & 56,143,121 & 56,609,104 & 57,872,099 & 58,070,077 & 58,324,067 & 64,823,801 & 65,426,701 & 65,725,421 & 65,794,806 & 66,079,791 & 66,355,245 & 66,570,526 \\
19 & 51,734,926 & 53,970,664 & 55,420,133 & 56,429,139 & 57,389,684 & 58,026,044 & 57,807,255 & 64,435,084 & 65,175,915 & 65,608,367 & 65,736,235 & 66,057,197 & 66,188,760 & 66,351,043 \\
20 & 52,066,233 & 53,844,254 & 55,306,358 & 56,168,419 & 56,834,780 & 57,322,399 & 57,930,955 & 64,026,114 & 64,482,213 & 64,845,907 & 65,324,118 & 65,663,196 & 65,946,061 & 65,795,199 \\
21 & 54,499,372 & 55,151,054 & 55,817,487 & 56,104,835 & 57,282,099 & 57,680,431 & 58,491,658 & 63,854,665 & 64,377,391 & 64,858,096 & 64,933,132 & 65,479,905 & 65,461,356 & 65,605,633 \\
22 & 53,002,234 & 53,689,249 & 54,689,980 & 55,709,870 & 56,393,115 & 56,890,668 & 57,461,711 & 63,953,011 & 64,210,926 & 64,748,681 & 64,882,441 & 65,007,446 & 65,115,475 & 65,219,898 \\
23 & 54,475,618 & 54,209,390 & 55,114,979 & 55,800,549 & 56,745,566 & 56,890,699 & 57,313,138 & 63,693,211 & 64,323,138 & 64,361,084 & 64,798,794 & 64,818,666 & 65,024,302 & 64,992,457 \\
24 & 52,696,844 & 53,885,907 & 54,088,030 & 55,057,040 & 55,463,131 & 56,208,303 & 56,444,958 & 63,400,561 & 63,820,771 & 63,700,022 & 64,438,090 & 64,258,752 & 64,141,101 & 64,186,891 \\
25 & 50,318,283 & 51,269,205 & 52,481,347 & 53,465,984 & 53,891,662 & 54,781,920 & 54,713,197 & 63,232,652 & 63,129,362 & 63,259,586 & 63,725,082 & 63,910,320 & 63,867,932 & 63,988,413 \\
26 & 50,229,788 & 51,354,758 & 52,362,605 & 53,242,078 & 53,893,195 & 54,326,699 & 54,530,116 & 61,826,182 & 61,893,288 & 62,274,584 & 62,752,607 & 62,746,811 & 62,887,121 & 63,116,564 \\
27 & 52,206,436 & 52,688,837 & 53,341,050 & 53,769,840 & 54,270,899 & 54,576,546 & 55,060,048 & 60,352,198 & 60,726,965 & 60,923,515 & 61,131,020 & 60,963,222 & 61,265,266 & 61,663,726 \\
28 & 49,682,706 & 50,249,019 & 51,876,101 & 52,495,005 & 52,542,959 & 53,202,973 & 53,174,103 & 60,534,641 & 60,288,939 & 60,079,857 & 59,975,267 & 60,096,463 & 59,981,804 & 60,079,474 \\
29 & 49,857,350 & 50,055,633 & 51,690,368 & 51,477,619 & 51,245,707 & 51,586,655 & 52,023,933 & 61,446,881 & 60,521,500 & 59,891,967 & 59,610,978 & 59,336,539 & 59,242,416 & 59,109,188 \\
30 & 48,055,193 & 47,429,451 & 48,033,530 & 48,151,423 & 48,174,882 & 48,183,182 & 48,215,800 & 59,390,999 & 57,149,284 & 56,340,637 & 55,831,719 & 55,327,710 & 54,997,345 & 54,953,972 \\
31 & 39,251,259 & 39,137,018 & 39,388,512 & 39,273,372 & 39,230,840 & 39,250,211 & 39,423,466 & 47,469,654 & 45,842,988 & 45,103,473 & 44,628,970 & 44,396,335 & 44,093,542 & 44,099,541 \\
Avg Context & 52,167,024 & 53,402,932 & 54,503,993 & 55,212,777 & 56,103,781 & 56,535,303 & 56,929,119 & 63,427,271 & 63,783,267 & 64,121,472 & 64,304,913 & 64,524,716 & 64,580,953 & 64,670,137 \\
32 & 50,487,094 & 49,193,302 & 48,748,380 & 48,353,375 & 48,143,528 & 47,788,231 & 47,682,812 & 73,626,535 & 69,054,469 & 66,707,831 & 64,778,405 & 63,608,111 & 62,438,327 & 62,279,482 \\
33 & 61,392,551 & 59,894,701 & 58,602,818 & 58,523,094 & 57,350,370 & 57,349,346 & 56,865,365 & 127,465,382 & 117,081,737 & 110,181,057 & 106,211,754 & 102,176,536 & 100,298,529 & 98,864,531 \\
34 & 61,328,405 & 61,880,320 & 61,897,331 & 61,603,725 & 61,037,948 & 60,910,761 & 61,148,637 & 183,840,986 & 178,535,599 & 174,318,924 & 169,430,877 & 164,785,135 & 162,605,626 & 161,446,058 \\
35 & 62,108,896 & 64,859,473 & 66,373,167 & 66,487,737 & 66,963,760 & 67,607,923 & 67,185,378 & 207,065,353 & 212,630,432 & 212,924,641 & 212,313,484 & 210,389,783 & 209,373,568 & 209,264,567 \\
36 & 64,491,225 & 66,128,068 & 68,573,633 & 68,479,538 & 69,624,085 & 69,468,930 & 70,076,197 & 107,042,066 & 102,720,927 & 98,446,681 & 95,544,431 & 91,884,559 & 89,921,371 & 89,176,573 \\
37 & 64,176,596 & 67,738,610 & 71,189,574 & 72,648,126 & 74,649,180 & 75,348,320 & 76,011,136 & 78,418,713 & 75,370,613 & 73,491,350 & 72,095,517 & 70,419,121 & 69,798,088 & 69,480,522 \\
38 & 83,023,501 & 86,268,700 & 92,242,626 & 95,541,179 & 100,137,087 & 103,141,517 & 104,996,137 & 75,331,828 & 73,633,103 & 72,372,719 & 71,451,162 & 70,344,087 & 69,870,019 & 69,555,251 \\
39 & 114,997,010 & 123,345,929 & 132,119,453 & 137,345,936 & 144,729,841 & 148,911,774 & 152,506,274 & 72,416,590 & 71,374,676 & 70,711,375 & 69,920,343 & 69,170,308 & 68,691,759 & 68,547,626 \\
40 & 72,923,797 & 75,060,147 & 76,865,835 & 77,568,398 & 78,566,453 & 79,311,885 & 80,082,621 & 73,411,371 & 72,849,906 & 71,665,061 & 71,318,529 & 70,698,319 & 70,306,675 & 70,168,524 \\
41 & 62,139,748 & 64,712,748 & 65,532,787 & 65,890,626 & 66,512,689 & 66,430,631 & 67,029,855 & 73,147,695 & 72,602,565 & 71,681,923 & 71,271,172 & 70,741,320 & 70,453,928 & 70,443,127 \\
42 & 61,518,278 & 63,119,965 & 63,331,920 & 63,642,209 & 64,136,809 & 63,861,430 & 63,990,245 & 72,618,045 & 71,781,987 & 71,440,631 & 71,063,281 & 70,644,161 & 70,394,990 & 70,299,681 \\
43 & 58,893,899 & 59,747,917 & 60,630,537 & 60,111,792 & 60,897,906 & 60,413,416 & 60,685,337 & 72,647,660 & 71,716,108 & 71,179,077 & 70,986,376 & 70,431,462 & 70,058,904 & 70,263,439 \\
44 & 57,229,696 & 57,312,370 & 57,433,177 & 58,104,220 & 57,890,592 & 58,313,878 & 58,439,482 & 72,211,146 & 71,314,701 & 70,977,086 & 70,897,382 & 70,417,594 & 70,322,615 & 70,408,343 \\
45 & 55,873,454 & 56,139,949 & 56,811,592 & 56,589,596 & 57,622,149 & 57,870,333 & 58,004,104 & 72,853,238 & 71,729,416 & 71,312,109 & 71,062,683 & 70,958,396 & 70,941,043 & 70,812,855 \\
46 & 54,966,654 & 55,910,428 & 56,508,715 & 56,557,283 & 57,408,451 & 57,383,704 & 58,025,074 & 73,234,042 & 72,148,832 & 71,787,141 & 71,388,367 & 71,441,543 & 70,955,811 & 71,090,486 \\
47 & 55,258,100 & 55,540,073 & 56,121,577 & 56,868,270 & 57,182,726 & 57,490,540 & 58,014,824 & 71,452,519 & 70,861,810 & 70,783,165 & 70,262,176 & 70,511,868 & 70,420,548 & 70,296,965 \\
48 & 53,806,444 & 54,635,210 & 54,818,474 & 55,121,630 & 55,692,510 & 56,391,570 & 56,482,717 & 65,034,377 & 65,452,587 & 65,749,667 & 66,046,086 & 66,391,820 & 66,416,952 & 66,424,818 \\
Avg Continuation & 64,389,138 & 65,969,877 & 67,517,741 & 68,202,161 & 69,326,240 & 69,882,011 & 70,425,070 & 92,459,856 & 90,638,792 & 89,160,614 & 88,002,472 & 86,765,537 & 86,074,632 & 85,813,109 \\
Avg Sentence & 56,495,689 & 57,853,725 & 59,113,028 & 59,813,184 & 60,786,735 & 61,262,262 & 61,708,935 & 73,709,645 & 73,294,599 & 72,989,502 & 72,697,798 & 72,401,674 & 72,193,298 & 72,158,273\\ 
\bottomrule
\end{tabular}%
}
\caption{2 Gram Statistics for half-memorized and quarter-memorized content.}
\label{tab:2 gram 2}
\end{table*}

\end{document}